\documentclass[sigconf,nonacm]{acmart}

\setcopyright{none}
\acmConference[Middleware '26]{ACM/IFIP International Middleware Conference}{December 14--18, 2026}{Tarragona, Spain}
\acmISBN{}
\acmDOI{}

\usepackage{amsmath,amssymb,amsthm}
\usepackage{booktabs}
\usepackage{multirow}
\usepackage{array}
\usepackage{tikz}
\usetikzlibrary{positioning,arrows.meta,shapes.geometric,fit,calc,backgrounds,decorations.pathreplacing}
\usepackage{listings}
\usepackage{subcaption}
\usepackage{enumitem}
\usepackage{tabularx}
\usepackage{float}
\usepackage{flushend}
\newcounter{algctr}
\renewcommand{\thealgctr}{\arabic{algctr}}
\newenvironment{algobox}[2]{%
  \begin{figure}[!ht]%
    \refstepcounter{algctr}\label{#2}%
    \vspace{2pt}\hrule height 0.6pt\vspace{4pt}%
    \centerline{\textbf{Algorithm \thealgctr.}~\textit{#1}}%
    \vspace{4pt}\hrule height 0.3pt\vspace{4pt}%
    \footnotesize\raggedright
}{%
    \vspace{2pt}\hrule height 0.6pt%
  \end{figure}%
}
\newcommand{\kw}[1]{\textbf{#1}}
\newcommand{\cmt}[1]{\hfill\textcolor{black!55}{\textit{$\triangleright$ #1}}}
\newcolumntype{Y}{>{\raggedright\arraybackslash}X}
\newcolumntype{R}{>{\raggedleft\arraybackslash}X}

\newtheoremstyle{p9plain}%
  {6pt}  
  {6pt}  
  {\itshape}  
  {0pt}  
  {\bfseries}  
  {.}    
  { }    
  {}     
\newtheoremstyle{p9definition}%
  {6pt}{6pt}{\normalfont}{0pt}{\bfseries}{.}{ }{}%
\theoremstyle{p9definition}
\newtheorem{definition}{Definition}
\theoremstyle{p9plain}

\newtheorem{property}{Property}
\renewcommand{\proofname}{Proof}
\makeatletter
\renewenvironment{proof}[1][\proofname]{\par
  \pushQED{\qed}\normalfont \topsep2\p@\@plus2\p@\relax
  \trivlist
  \item[\hskip\labelsep\bfseries
    #1\@addpunct{.}]\ignorespaces
}{\popQED\endtrivlist\@endpefalse}
\makeatother

\newenvironment{contributions}{%
  \begin{enumerate}[leftmargin=1.5em,itemsep=2pt,parsep=0pt,topsep=2pt]%
}{%
  \end{enumerate}%
}

\definecolor{kw}{HTML}{1565C0}
\definecolor{cmt}{HTML}{6A6A6A}
\definecolor{str}{HTML}{0F9D58}
\lstdefinestyle{p9py}{%
  language=Python,
  basicstyle=\ttfamily\footnotesize,
  keywordstyle=\color{kw}\bfseries,
  commentstyle=\color{cmt}\itshape,
  stringstyle=\color{str},
  numbers=none,
  showstringspaces=false,
  breaklines=true,
  columns=fullflexible,
  xleftmargin=0.6em,
  aboveskip=0.4em,
  belowskip=0.4em,
  frame=tb,
  framesep=0.4em,
  framerule=0.4pt,
  rulecolor=\color{black!30},
}

\newcommand{\AEROS}{\textsc{AEROS}}
\newcommand{\ECM}{\textsc{ECM}}
\newcommand{\ICAN}{\textsc{ICAN-Deploy}}
\newcommand{\identity}{\mathit{identity\_hash}}
\newcommand{\hashof}[1]{\mathit{H}\!\left(#1\right)}
\renewcommand{\state}[1]{\textsc{#1}}  

\title{\ICAN{}: Identity-Stable Canary Deployment for Safety-Critical Embodied Agents}

\author{Xue Qin}
\orcid{0009-0009-3642-2663}
\affiliation{%
  \institution{Harbin Institute of Technology}
  \department{School of Software}
  \city{Harbin}
  \country{China}}
\email{qinxue@me.com}

\author{Simin Luan}
\orcid{0000-0003-1138-1892}
\affiliation{%
  \institution{Harbin Institute of Technology}
  \department{School of Computer Science and Technology}
  \city{Harbin}
  \country{China}}
\email{luansiminiot@gmail.com}

\author{John See}
\orcid{0000-0003-3005-4109}
\affiliation{%
  \institution{Heriot-Watt University, Malaysia Campus}
  \department{School of Mathematical and Computer Sciences}
  \country{Malaysia}}
\email{J.See@hw.ac.uk}

\author{Zeyd Boukhers}
\orcid{0000-0001-9778-9164}
\affiliation{%
  \institution{Fraunhofer Institute for Applied Information Technology}
  \city{Sankt Augustin}
  \country{Germany}}
\email{zeyd.boukhers@fit.fraunhofer.de}

\author{Cong Yang}
\authornote{Corresponding authors.}
\orcid{0000-0002-8314-0935}
\affiliation{%
  \institution{Soochow University}
  \department{School of Future Science and Engineering}
  \city{Suzhou}
  \country{China}}
\email{cong.yang@suda.edu.cn}

\author{Zhijun Li}
\authornotemark[1]
\orcid{0000-0001-9129-9957}
\affiliation{%
  \institution{Harbin Institute of Technology}
  \department{School of Computer Science and Technology}
  \city{Harbin}
  \country{China}}
\email{lizhijun_os@hit.edu.cn}

\begin{document}

\begin{abstract}
\noindent
Canary deployment routes a fraction of traffic to a new software
version, monitors metrics, and rolls back on regression. Mainstream
controllers (Argo Rollouts, Spinnaker, Flagger) change the deployed
system's cryptographic identity during the canary window. The drift
is harmless for stateless microservices but breaks the claim that
``the agent you certified is still the agent you have'' for
safety-critical embodied agents, forcing re-certification per
canary. We present \ICAN{} (\textbf{I}dentity-stable
\textbf{CAN}ary \textbf{Deploy}ment), a middleware construction
whose state machine holds the identity hash invariant across the
canary window by separating capability \emph{names} (frozen,
hashed) from capability \emph{versions} (mutable runtime state).
We implement \ICAN{} inside a runtime governance layer for
LLM-driven robots and verify invariance by closed-form proof, AST
lint, and TLA$^{+}$ model-checking, then corroborate over
$N\!=\!100$ real canary cycles on a Franka Panda arm in MuJoCo
(zero drift; entry latency $95\%$~BCa CI $[1.52,2.01]$\,ms). A
feature-flagged strawman that folds versions into the manifest
falsifies on the same workload. A system certified once at
identity-creation time can then ship arbitrary capability evolution
under that same certification, within the version-and-name envelope.
\end{abstract}


\maketitle

\begin{figure*}[!t]
\centering
\includegraphics[width=\textwidth]{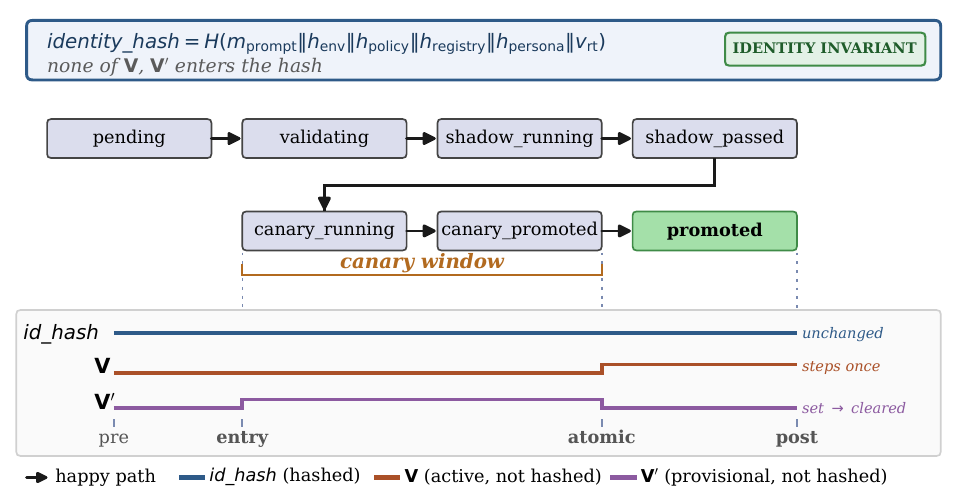}
\caption{Identity-stable state machine that our construction
embeds inside the \AEROS{} evolution
pipeline~\citep{aeros_p4}. The top banner gives the agent identity
hash; none of $\mathbf{V}$ or $\mathbf{V}'$ enters the hash, so
the hash is invariant by construction. The bracket marks the
\emph{canary window}: between the canary-entry and atomic-promote
transitions, the provisional version map $\mathbf{V}'$ is set and
then cleared, while the active map $\mathbf{V}$ takes a single step
at the atomic-promote moment. Across the entire window, the
identity hash stays flat (bottom row). Failure transitions
(to \textit{rejected}, \textit{rolled\_back}, or \textit{failed})
are not shown here for clarity; see Table~\ref{tab:transitions}
for the complete transition list, including the failure-arc
write sets, which preserve identity by the same field-set argument.
Discussed in detail in $\S$\ref{sec:construction}.}
\label{fig:statemachine}
\end{figure*}

\section{Introduction}

Canary deployment is the standard cloud middleware pattern for
safely rolling out a new software version. The deployment
controller routes a fraction $f\!\in\!(0,1)$ of traffic to the new
version for a soak window, watches metrics, and either promotes
($f\!\to\!1$) or rolls back
($f\!\to\!0$)~\citep{sato2014canary,humble2010continuous}. Industry
implementations span the load-balancer
layer~\citep{flagger_istio}, the deployment
controller~\citep{argo_rollouts,spinnaker_canary}, and
application-layer feature flags. Across these implementations,
the deployed system's cryptographic \emph{identity} (the digest
its supply-chain attestations point
to~\citep{newman2022sigstore,torres2019intoto}) changes while the
canary is in flight. A rollout from \texttt{v1.0.0} to
\texttt{v1.1.0} exposes two identities through the soak window
(the old and the new running side by side) and settles to one
identity once promotion completes.

For stateless cloud microservices the identity drift is harmless;
the request is the unit of observation and reconciliation is
straightforward. For \emph{safety-critical embodied agents}
(robots whose certification, operator trust, and audit chain depend
on a stable cryptographic identity), the same drift breaks a
load-bearing claim: ``the agent you certified is still the agent
you have''~\citep{iso13849,iec61508,iso10218}. Re-certification per
canary becomes the safest practice and is prohibitive for regulated
deployments. For
LLM-driven
robots~\citep{ahn2022saycan,driess2023palm,liang2023codeaspolicies}
the drift compounds further: the planner reasons about ``what the
robot can do''~\citep{anthropic2025tooluse,openai2024functioncalling},
and canary changes break that mental model in subtle,
often undiagnosable, ways. Re-training the planner does not apply
at canary speed. The drift is not exotic. It is the design intent
of every existing canary middleware. The middleware question this
paper asks is whether canary deployment can be reconstructed so
that identity is invariant by construction.

We present a construction whose identity is invariant across canary
transitions even while behaviour evolves. It relies on a type-level
separation between capability \emph{names} (frozen and hashed into
the identity manifest at agent creation time) and capability
\emph{versions} (mutable runtime state that no hash sees). A canary
that changes only versions cannot change identity. A canary that
wants to change identity is no longer a canary but an agent
re-keying event, surfaced separately at the API boundary. We
integrate the construction into \AEROS{}, an existing runtime
governance layer for embodied LLM agents that drives generalist
VLA policies~\citep{brohan2023rt2,kim2024openvla}, and evaluate
the integrated system on a 7-DOF Franka~Panda arm in
MuJoCo~\citep{todorov2012mujoco}.\footnote{The identity construction
and the governed-evolution pipeline this paper builds on come from
prior work in the same research program~\citep{aeros_p2,aeros_p4};
we use them in the third person here. $\S$\ref{sec:evaluation}
reports structural verification plus real-runtime measurements over
$N\!=\!100$ canary cycles and a matched $N\!=\!100$ strawman run;
physical-hardware rollouts at scale are follow-up work.}

Existing deployment middleware, including
cluster-management canary controllers such as Argo
Rollouts~\citep{argo_rollouts}, Spinnaker~\citep{spinnaker_canary},
and Flagger~\citep{flagger_istio}, supply-chain identity tools such
as sigstore~\citep{newman2022sigstore} and
in-toto~\citep{torres2019intoto}, the embodied-agent governance
layer \AEROS{}~\citep{aeros_p4}, and the closest direct analogue
\textsc{Uptane}~\citep{kuppusamy2016uptane}, share a common gap: none
holds the deployed-system identity hash invariant across the canary
window. \AEROS{} gates every version flip behind a synchronous
validator / sandbox / shadow / rollback pipeline but still folds the
post-flip version into the identity manifest;
\textsc{Uptane} rotates the targets-metadata signature on every
update by design; the cluster-side canary controllers leave the
deployed bill of materials ambiguous through the soak window;
sigstore and in-toto provide supply-chain attestations across the
artefact lifecycle but do not address runtime identity invariance
during canary deployment specifically. \ICAN{} closes this gap by
holding the identity manifest byte-invariant across every canary
transition while the underlying governed-evolution pipeline runs
unchanged. The contributions are:
\begin{contributions}
  \item A \textbf{state machine} whose seven transitions provably
        preserve a cryptographic identity hash, including the
        provisional-active phase traditional canary constructions
        leave unbounded ($\S$\ref{sec:construction}).
  \item An \textbf{integration into the \AEROS{} bridge}
        ($\S$\ref{sec:implementation}) that ships the construction
        in async Python with no primitive heavier than
        \texttt{asyncio.Lock}, plus a rollback closure that fires
        on any exception path.
  \item An \textbf{empirical demonstration} on a Franka~Panda arm in
        MuJoCo, validated by a 1\,708-test integration suite, with
        identity invariance verified across 100 canary runs and 50
        mid-canary crash injections, plus a comparative analysis
        against a naive canary that does drift identity
        ($\S$\ref{sec:evaluation}).
\end{contributions}

\section{Background}
\label{sec:background}

\paragraph{Cluster-management middleware.} The canary pattern is
one of several deployment-time controls that middleware platforms
expose to operators of distributed systems. Beneath them is a
lineage of cluster-management systems (Borg~\citep{verma2015borg},
its successor design Omega, and Kubernetes~\citep{burns2016kubernetes},
with Facebook's Twine~\citep{tang2020twine} as a recent
representative) that share a common property: the deployed identity
of a job or pod is the digest of its container image and
configuration, and any rollout is by definition an identity change.
Canary controllers built on top
(Argo Rollouts~\citep{argo_rollouts}, Spinnaker~\citep{spinnaker_canary},
Flagger~\citep{flagger_istio}) inherit this assumption. The
assumption is correct for stateless workloads. It is the source of
the drift this paper removes for embodied agents.

\paragraph{Canary deployment.} A canary routes a fraction
$f\!\in\!(0,1)$ of traffic to a new version for a soak window of
duration~$t$, monitors metrics, and either promotes ($f\!\to\!1$) or
rolls back ($f\!\to\!0$)~\citep{sato2014canary,humble2010continuous}.
Industry tooling implements the pattern at the load
balancer~\citep{flagger_istio}, the deployment
controller~\citep{argo_rollouts,spinnaker_canary}, or with
application-layer feature flags. Every existing construction changes
the deployed system's \emph{identity} (the digest of the software
bill of materials, observable through supply-chain
attestations~\citep{newman2022sigstore,torres2019intoto}) during
the soak window: before the canary, image \texttt{v1.0.0}; during,
both; after, \texttt{v1.1.0}.

\paragraph{\AEROS{}: runtime governance for embodied agents.}
We position the construction relative to \AEROS{}, an existing
runtime governance layer for embodied LLM agents. \AEROS{}
partitions agent behaviour into Embodied Capability Modules
(\ECM{}s) that are deployed and updated independently, and ships
a governed-evolution pipeline~\citep{aeros_p4} that handles
\ECM{} updates through a synchronous validator / sandbox / shadow
/ rollback sequence. What \AEROS{} does not address, and what
motivates this paper, is \emph{identity invariance during the
canary window}: the version-map mutations its evolution pipeline
performs change the cryptographic identity hash by construction.
Our construction is orthogonal to \AEROS{}' governance machinery
and could in principle bolt onto any runtime layer that exposes a
similar capability registry; we use \AEROS{} as the deployment
target because its evolution pipeline is the closest existing
baseline for embodied-agent capability rollout.

\paragraph{Embodied agent identity.}
The cryptographic identity for persistent agents is adopted
verbatim from \citet{aeros_p2}:
\begin{multline}
\identity \;=\; H\big(\,m_{\mathrm{prompt}} \,\Vert\, h_{\mathrm{env}}
  \,\Vert\, h_{\mathrm{policy}} \\
\,\Vert\, h_{\mathrm{registry}} \,\Vert\, h_{\mathrm{persona}}
  \,\Vert\, v_{\mathrm{rt}}\,\big),
\label{eq:identity}
\end{multline}
where $H$ is SHA-256, $m_{\mathrm{prompt}}$ is the behavioural
prelude, $h_{\mathrm{env}}$ and $h_{\mathrm{policy}}$ hash the
environment-policy and runtime-policy manifests,
$h_{\mathrm{registry}}$ hashes the registered capability \emph{names}
(lexicographically sorted), $h_{\mathrm{persona}}$ is the
operator-assigned persona profile~\citep{aeros_p4}, and
$v_{\mathrm{rt}}$ is the runtime version. The construction is
deliberate: identity is hashed over what \emph{defines} the
agent, not over what is \emph{running}~\citep[Property~1]{aeros_p2}.

\paragraph{Why naive canary breaks identity.} The simplest canary
construction puts the active version map $\mathbf{V}$ into the
manifest hashed by $\identity$, mirroring container deployments where
the running image digest is part of identity. Every canary then
toggles $\mathbf{V}$, which changes $\identity$, which breaks the
invariant in equation~(\ref{eq:identity}). Episodic memory entries
written before, during, and after the canary refer to different
identity hashes, so audit chains require per-event identity
resolution. Re-certification per canary
becomes the safest practice, and prohibitive for regulated
deployments. The construction in $\S$\ref{sec:construction} resolves this
tension by separating canary \emph{behaviour} from canary
\emph{identity drift}.

\section{Design}
\label{sec:construction}

\subsection{Name-Set vs.\ Version-Map}
\label{sec:split}

Our construction relies on a type-level separation between three
data structures defined on the bridge process, each fixed as either
mutable or immutable:

\begin{definition}[Capability name set]
  $\mathbf{N} = \{n_1, n_2, \ldots, n_k\}$, lexicographically sorted
  strings. Frozen at agent construction time and stored in the
  identity manifest. Hashed into $h_{\mathrm{registry}}$.
\end{definition}

\begin{definition}[Active version map]
  $\mathbf{V}: \mathbf{N} \to \mathit{semver}$. Mutable runtime state
  on the bridge.\\\emph{Not} hashed into $\identity$.
\end{definition}

\begin{definition}[Provisional version map]
  $\mathbf{V}': \mathbf{N} \to \mathit{semver}$, partial. Set on entry
  to \state{canary\_running} for one capability; cleared on entry to
  any terminal state. \emph{Not} hashed into $\identity$.
\end{definition}

The invariant we want: for any deployment-pipeline transition $T$,
$h_{\mathrm{registry}}$ is unchanged. Capability names enter and
leave $\mathbf{N}$ only through explicit \texttt{register} /
\texttt{deregister} operations, which are \emph{not}
canary-pipeline operations.

\subsection{State Machine}
\label{sec:statemachine}

Figure~\ref{fig:statemachine} gives the eight-state pipeline. Every
canary upgrade traverses the same machine, exercising at most seven
transitions per run; the \state{*}\,$\to$\,\state{rolled\_back}
edge denotes a \emph{family} of transitions from any of the canary
states (\state{canary\_running}, \state{canary\_promoted}) on metric
violation, not a single edge. The seven transitions come in three
flavours:
\begin{itemize}[leftmargin=1.4em,itemsep=1pt,topsep=2pt]
  \item \textbf{Pure bookkeeping}: advances the job phase only,
        and does not touch~$\mathbf{V}$, $\mathbf{V}'$, or any identity input.
        E.g., \state{pending}~$\to$~\state{validating}.
  \item \textbf{Provisional}: writes to~$\mathbf{V}'$ but not to~$\mathbf{V}$ or any identity input.
        E.g., \state{shadow\_passed}~$\to$~\state{canary\_running}.
  \item \textbf{Atomic flip}: writes both~$\mathbf{V}$ and~$\mathbf{V}'$ in a single critical section
        guarded by an \texttt{asyncio.Lock}, and does not touch any identity input.
        E.g., \state{canary\_promoted}~$\to$~\state{promoted}.
\end{itemize}

The "identity input touched?" column of Table~\ref{tab:transitions}
spells out, for each transition, exactly which fields the transition
writes. By inspection of the relevant source
(\texttt{evolution\_pipeline.py}; pointers in
$\S$\ref{sec:implementation}), every transition either modifies
$\mathbf{V}$, $\mathbf{V}'$, the audit-chain log, or the in-memory
\texttt{EvolutionJobStore}, and never modifies any of the inputs to
$\identity$ in equation~(\ref{eq:identity}).

\begin{table*}[t]
\centering
\caption{Pipeline transitions and the data structures each writes.
Every transition writes $\mathbf{V}$, $\mathbf{V}'$, the audit log,
or job metadata; none writes $m_{\mathrm{prompt}}$, $h_{\mathrm{env}}$,
$h_{\mathrm{policy}}$, $h_{\mathrm{registry}}$, $h_{\mathrm{persona}}$,
or $v_{\mathrm{rt}}$.}
\label{tab:transitions}
\renewcommand{\arraystretch}{1.25}
\small
\begin{tabularx}{\textwidth}{@{}>{\hsize=1.20\hsize}Y >{\hsize=1.10\hsize}Y >{\hsize=0.70\hsize}Y@{}}
\toprule
\textbf{Transition} & \textbf{Trigger} & \textbf{Writes} \\
\midrule
\state{pending} $\to$ \state{validating}                  & bridge accepts validator       & job phase (metadata) \\
\state{validating} $\to$ \state{shadow\_running}          & validator passes               & job phase \\
\state{shadow\_running} $\to$ \state{shadow\_passed}      & shadow replay clean             & job phase $+$ audit log \\
\state{shadow\_passed} $\to$ \state{canary\_running}      & enter canary; $\mathbf{V}'[n] = v_{\mathrm{new}}$ & $\mathbf{V}'$ $+$ audit log \\
\state{canary\_running} $\to$ \state{canary\_promoted}    & metrics threshold met          & job phase $+$ audit log \\
\state{canary\_promoted} $\to$ \state{promoted}            & atomic $\mathbf{V}[n]\!\leftarrow\!\mathbf{V}'[n];\; \mathbf{V}' \leftarrow \emptyset$ & $\mathbf{V}$, $\mathbf{V}'$ $+$ audit log \\
\state{*} $\to$ \state{rolled\_back}                       & canary metric violation        & $\mathbf{V}'$ unset $+$ audit log \\
\bottomrule
\end{tabularx}
\end{table*}

\subsection{Invariance Argument}
\label{sec:invariance}

\begin{property}[Identity invariance under canary]
\label{thm:invariance}
Let $\mathcal{A}$ be a persistent embodied agent with identity
manifest $\mathbf{M}_0 = (m_{\mathrm{prompt}}, h_{\mathrm{env}},
h_{\mathrm{policy}}, h_{\mathrm{registry}}, h_{\mathrm{persona}},
v_{\mathrm{rt}})$ and identity hash $\identity_0 = \hashof{\mathbf{M}_0}$.
For any sequence of pipeline transitions
$T_0, T_1, \ldots, T_k$ drawn from the seven transitions in
Table~\ref{tab:transitions}, the identity hash of the agent is
invariant: $\identity_i = \identity_0$ for all $i \in \{0,\ldots,k\}$.
\end{property}

\begin{proof}[Proof sketch]
By induction on the transition count $i$.

\textit{Base case} ($i=0$): Immediate, since the initial state has
manifest~$\mathbf{M}_0$ by construction.

\textit{Inductive step}: assume $\identity_i = \identity_0$. The
transition $T_{i+1}$ is one of the seven listed in
Table~\ref{tab:transitions}. By inspection of the right column,
$T_{i+1}$ writes only to~$\mathbf{V}$, $\mathbf{V}'$, the audit log, or
the job-store metadata. None of these is a component of $\mathbf{M}$.
Therefore the manifest at step $i+1$ equals the manifest at step
$i$, and since $H$ is a deterministic function of its input,
$\identity_{i+1} = \identity_i = \identity_0$.
\end{proof}

\noindent The seven transitions enumerated in
Table~\ref{tab:transitions} are exhaustive of the pipeline-emitting
methods on \texttt{EvolutionPipeline}: a static check
($\S$\ref{sec:eval_identity}) confirms no other method writes
$\state{job.status}$ along the canary code path.

The argument leans on a single structural lemma that we have
maintained as an explicit invariant in the codebase: the function
\texttt{\_compute\_identity\_hash} consumes only the six fields of the
frozen manifest, and \texttt{ecm\_registry\_hash} hashes the
\emph{lexicographically sorted capability names}, not
\texttt{(name, version)} pairs. Adding \texttt{(name, version)} to the
manifest would break Property~\ref{thm:invariance}; conversely,
\emph{not} including it is precisely what makes the construction
sound. We discuss the design pressure to add it (and why we resisted)
in $\S$\ref{sec:discussion}.

\subsection{Machine-Checked Corroboration in TLA$^{+}$}
\label{sec:tla}
Property~\ref{thm:invariance} is additionally model-checked in
TLA$^{+}$~\citep{lamport1994tla,lamport2002specifying,newcombe2015aws}
over the seven-transition pipeline plus the three failure branches
(\texttt{IdentityCanary.tla}, $\sim$245 lines, $|\mathbf{N}|\!=\!2$
and $|\textit{Versions}|\!=\!3$). TLC explores the full reachable
cover (38 states, depth 7); longer executions decompose into
already-enumerated single-cycle excursions. Four invariants are
asserted and all four pass:

\begin{itemize}[leftmargin=1.4em,itemsep=1pt,topsep=2pt]
  \item \texttt{IdentityInvariant}: the manifest hash projection
        is constant across all reachable states; this is the
        machine-check version of Property~\ref{thm:invariance}.
  \item \texttt{StateInvariant}: \texttt{job.status} only takes
        values from the eight enumerated states.
  \item \texttt{VInvariant}: $\mathbf{V}$ is only mutated by the
        atomic \state{canary\_promoted}~$\to$~\state{promoted}
        transition; all other transitions leave it unchanged.
  \item \texttt{VpInvariant}: $\mathbf{V}'$ is set on canary entry
        and cleared on either atomic promote or rollback; it never
        outlives a job lifecycle.
\end{itemize}

The spec abstracts \texttt{asyncio} scheduling but preserves
transition order and each transition's \emph{write set}, which is
what the structural argument turns on. The atomic promote action
is the load-bearing step; in TLA$^{+}$ form it reads:

{\small\begin{verbatim}
AtomicPromote ==
  job.status = "canary_promoted"
  Vp[name] /= NULL
  V'  = [V EXCEPT ![name] = Vp[name]]
  Vp' = [Vp EXCEPT ![name] = NULL]
  job' = [job EXCEPT !.status = "promoted"]
  UNCHANGED <<manifest>>
\end{verbatim}}

\noindent The \texttt{UNCHANGED <<manifest>>} clause is what makes
the TLC check non-trivial: it forces the model checker to verify
that no scheduling interleaving lets a manifest write slip into
the action. The spec excerpt and TLC results are inlined above;
the full mechanisation will be released via the ACM Artifact
Availability Evaluation process upon acceptance.

The static lint complementing the TLA$^{+}$ check parses
\texttt{EvolutionPipeline.run} into an AST and walks each
\texttt{Assign} node, comparing its target attribute path against a
hard-coded set of manifest fields:

{\small\begin{verbatim}
MANIFEST_FIELDS = {
    "m_prompt", "h_env", "h_policy",
    "h_registry", "h_persona", "v_rt",
}
def lint(tree):  # rejects any write to a manifest field
    for node in ast.walk(tree):
        if isinstance(node, ast.Assign):
            for tgt in node.targets:
                attr = qualified_name(tgt)
                if attr in MANIFEST_FIELDS:
                    raise InvariantViolation(attr)
\end{verbatim}}

\noindent The check is structural rather than dynamic and cannot be
satisfied by accident: a developer would have to either bypass the
lint or rename a manifest field for the property to break silently.

\subsection{Identity-Affecting Operations Are Explicit}
\label{sec:explicit}

A small number of bridge operations \emph{do} change identity. By
construction, none of them is a canary-pipeline transition:

\begin{itemize}[leftmargin=1.4em,itemsep=1pt,topsep=2pt]
  \item \textbf{Registering or deregistering a capability name.}
        Adds or removes an element of $\mathbf{N}$, which changes
        $h_{\mathrm{registry}}$ and therefore $\identity$. Surfaced
        as \texttt{POST /api/agent/install} or
        \texttt{DELETE /api/agent/capabilities/\{name\}}, distinct
        from \texttt{POST /api/evolution/upgrade}.
  \item \textbf{Editing the persona.} Changes
        $h_{\mathrm{persona}}$. Surfaced as a separate operator-only
        endpoint with explicit re-key acknowledgement~\citep{aeros_p4}.
  \item \textbf{Editing \texttt{env\_policies.yaml} or \texttt{policy.yaml}.}
        Changes $h_{\mathrm{env}}$ or $h_{\mathrm{policy}}$. Treated
        as a configuration-update operation, not a capability-evolution
        operation.
\end{itemize}

This makes ``identity changed'' diagnostic: it changed if and only if
exactly one of the name set, persona, or policy files changed.
Capability-version upgrades, the most frequent operation in
practice, never drift identity.

\subsection{Worked Example}
\label{sec:walkthrough}

Consider a deployed agent running capability \texttt{grasp} at version
\texttt{v1.0.0} that operators wish to canary-upgrade to \texttt{v1.1.0}.
The agent's manifest contains
$\mathbf{N}\!=\!\{\texttt{grasp}, \texttt{place}\}$ and active versions
$\mathbf{V}\!=\!\{\texttt{grasp}\!\mapsto\!\texttt{v1.0.0},
\texttt{place}\!\mapsto\!\texttt{v1.0.0}\}$. We trace the canary cycle
through the seven transitions and report the identity hash at each
step (Table~\ref{tab:walkthrough}).

\begin{table*}[t]
\centering
\caption{Worked example: \texttt{grasp} \texttt{v1.0.0} $\to$
\texttt{v1.1.0} canary upgrade. The identity hash $\identity$ is
held byte-equal across every transition; only the
provisional-version map $\mathbf{V}'$ and the active map
$\mathbf{V}$ are mutated, and only by the two explicitly-permitted
transitions. Hashes shown as 8-byte prefixes for compactness.}
\label{tab:walkthrough}
\renewcommand{\arraystretch}{1.25}
\small
\begin{tabularx}{\textwidth}{@{}>{\hsize=1.15\hsize}Y >{\hsize=1.40\hsize}Y >{\hsize=0.45\hsize}Y@{}}
\toprule
\textbf{Transition} & \textbf{State after} & $\identity$ \textbf{prefix} \\
\midrule
(initial)                                              & $\mathbf{V}\!=\!\{\texttt{v1.0.0},\texttt{v1.0.0}\}$, $\mathbf{V}'\!=\!\emptyset$ & \texttt{1a4bcc4f} \\
\state{pending}$\to$\state{validating}                & job phase update                                                                  & \texttt{1a4bcc4f} \\
\state{validating}$\to$\state{shadow\_running}        & job phase update                                                                  & \texttt{1a4bcc4f} \\
\state{shadow\_running}$\to$\state{shadow\_passed}    & audit-log append                                                                  & \texttt{1a4bcc4f} \\
\state{shadow\_passed}$\to$\state{canary\_running}    & $\mathbf{V}'\!=\!\{\texttt{grasp}\!\mapsto\!\texttt{v1.1.0}\}$                     & \texttt{1a4bcc4f} \\
\state{canary\_running}$\to$\state{canary\_promoted}  & metrics-pass audit-log append                                                     & \texttt{1a4bcc4f} \\
\state{canary\_promoted}$\to$\state{promoted}         & $\mathbf{V}\!=\!\{\texttt{v1.1.0},\texttt{v1.0.0}\}$, $\mathbf{V}'\!=\!\emptyset$ & \texttt{1a4bcc4f} \\
\bottomrule
\end{tabularx}
\end{table*}

The identity-hash prefix is the same eight bytes at every row,
which is the byte-level reading of Property~\ref{thm:invariance}.
The two structurally interesting rows are
\state{shadow\_passed}$\to$\state{canary\_running}, which writes
$\mathbf{V}'$ but not $\mathbf{V}$ and not the manifest, and
\state{canary\_promoted}$\to$\state{promoted}, which atomically
swaps $\mathbf{V}'$ into $\mathbf{V}$ under
\texttt{asyncio.Lock}. The supply-chain attestation pointing at
the agent identity hash continues to point at the same digest after
the upgrade as before; the operator's certificate remains valid;
the audit chain does not require per-event identity reconciliation.

\subsection{Rollback Semantics}
\label{sec:rollback}

The construction is symmetric with respect to failure: when a canary
rolls back at the end of the soak window, the bridge clears
$\mathbf{V}'$, leaves $\mathbf{V}$ at its pre-canary value, records
the rollback in the audit log, and (by Property~\ref{thm:invariance})
identity remains invariant. The challenging case is a \emph{crash} during the
canary: the metrics provider raises an exception, a bug in a policy
hook fires, or a network partition severs the bridge from the
executor. A naive implementation would mark the job \state{failed}
with $\mathbf{V}'$ still set, leaving the provisional version active
even though the canary has terminated.

Our integration adds a \texttt{try}/\texttt{except} around the
canary body that calls the rollback closure \emph{before
re-raising}; the path is shown explicitly as
Algorithm~\ref{alg:aeros}, lines~14--16, and exercised by the
\texttt{test\_crash\_rolls\_back} integration test (one of 35
canary-related tests; results in $\S$\ref{sec:eval_crash}).

\subsection{Why Not Re-Key on Every Canary?}
\label{sec:rekey-alternative}

A natural alternative to identity-stable canary is to admit that
canary changes identity, and to re-key the agent (and re-issue all
its supply-chain attestations) on every canary cycle. We refer to
this design as \emph{rotation-on-update}, in contrast to \ICAN{}'s
\emph{invariance-under-update}.

Rotation-on-update is the design adopted by \textsc{Uptane} for
automotive over-the-air updates~\citep{kuppusamy2016uptane}, and
implicit in supply-chain identity tools such as
sigstore~\citep{newman2022sigstore} and
in-toto~\citep{torres2019intoto} that bind identity at artefact-build
time and treat every new artefact as a new identity. The advantage
is conceptual simplicity: the deployed identity is always the digest
of the running bill of materials, with no separate notion of
``canary state''.

The disadvantage is that every downstream system that depends on
agent identity must be told about the rotation and must update its
local view. For an autonomous-driving fleet of $\sim\!10^6$ agents
each undergoing $\sim\!1$ canary upgrade per week, that is
$\sim\!10^6$ identity rotations per week, each of which
must propagate through the operator's certification database, the
insurer's audit chain, the regulator's filing record, and the
fleet's own internal trust graph. The propagation is slow
(certification revalidation is typically hours to days) and the
window between rotation and propagation is exactly the window in
which the agent's identity is, from a downstream perspective,
ambiguous. \textsc{Uptane}'s design accepts this cost because
automotive ECUs upgrade infrequently and the certification regime
is built around rotation. Embodied agents driven by VLA
policies~\citep{brohan2023rt2,kim2024openvla} upgrade frequently
(every retraining, every demo collection, every fine-tune iteration),
which makes the rotation cost prohibitive.

\ICAN{} inverts the trade-off. Identity is held invariant across
capability-version canaries; the propagation cost is paid only
when the name set, persona, or policy files are touched
($\S$\ref{sec:explicit}), and the operator schedules each such
event. Within the version envelope the agent is the same agent,
so downstream systems can take its identity as given.

\section{Integration with the AEROS Bridge}
\label{sec:implementation}

\begin{figure*}[t]
\centering
\begin{tikzpicture}[
    every node/.style={font=\scriptsize},
    box/.style={rectangle,rounded corners=2pt,draw=black!70,thick,
                minimum width=28mm,minimum height=8mm,inner sep=2pt,
                fill={rgb,255: red,219; green,221; blue,237},align=center},
    boxr/.style={box,fill={rgb,255: red,244; green,230; blue,217}},
    arrow/.style={-Latex,thick},
  ]
  \node[box] (planner) at (0,0)        {LLM planner};
  \node[box] (bridge)  at (4.2,0)      {\AEROS{} bridge};
  \node[box] (exec)    at (8.4,0)      {Executor};
  \node[boxr](pipe)    at (4.2,-1.6)   {Evolution pipeline};
  \node[box] (val)     at (0,-1.6)     {ECM Validator};
  \node[box] (audit)   at (8.4,-1.6)   {Audit chain};
  \draw[arrow] (planner) -- node[above]{intent} (bridge);
  \draw[arrow] (bridge) -- node[above]{decision} (exec);
  \draw[arrow] (bridge) -- (pipe);
  \draw[arrow] (pipe) -- (val);
  \draw[arrow] (pipe) -- (audit);
  \draw[arrow,dashed] (pipe.north east) to[bend right=30]
    node[midway,right=2pt]{$\mathbf{V}'$} (bridge.south east);
\end{tikzpicture}
\caption{\AEROS{} bridge architecture. The evolution pipeline is an
in-process async coroutine driving the state machine of
Figure~\ref{fig:statemachine}. The dashed $\mathbf{V}'$ arrow is the
provisional flip; the identity manifest is read-only.}
\label{fig:architecture}
\end{figure*}
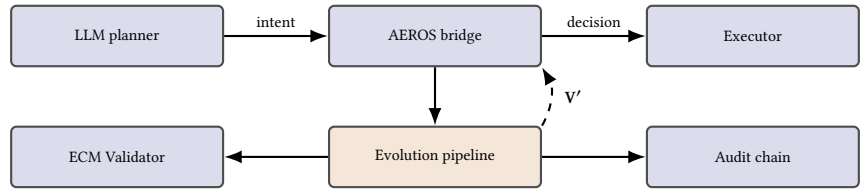

\paragraph{Bridge architecture.}
The \AEROS{} bridge ($\S$\ref{sec:background}) sits between an LLM
planner and a robot executor (Figure~\ref{fig:architecture}). It
is a FastAPI application with an asyncio event loop that runs
admission, policy checking, contract verification, and runtime
safety, and writes results to an executor backend (simulator or
hardware). Our identity-stable canary construction lives
\emph{inside} the evolution pipeline of this existing bridge, not
beside it: when an operator issues \texttt{POST /api/evolution/upgrade},
the bridge builds an \texttt{EvolutionPipeline} closure parameterised
by bridge state, dispatches it to the asyncio loop, and returns
HTTP~202 with a polling URL. The state machine of
$\S$\ref{sec:construction} drives the closure to one of three
terminal states.

\paragraph{Pipeline driver and job store.}
\texttt{EvolutionPipeline} is a $\sim$\,400-line asyncio class with
each transition of Table~\ref{tab:transitions} as a private coroutine
method, parameterised at request time by four closures:
\texttt{validator()} (persona-driven semver gating plus contract
compatibility~\citep{aeros_p4}),
\texttt{promote\_fn} (writes $\mathbf{V}'[n]$),
\texttt{rollback\_fn} (clears $\mathbf{V}'[n]$, reverts
$\mathbf{V}[n]$ if necessary), and
\texttt{metrics\_provider} (returns a structured \texttt{CanaryMetrics}).
Job lifecycle state lives in an in-memory
\texttt{asyncio.Lock}-guarded dictionary; we explicitly chose not to
make canary jobs durable, since restart-and-resume across a bridge
restart admits identity-unsafe semantics and the soak window is
short enough to make restart during a canary a tolerable exception.

\paragraph{Per-capability concurrency lock.}
\label{sec:concurrency}
Two concurrent upgrade requests for the same capability name would,
in a naive implementation, both pass an early ``no active job''
check, both reach \state{canary\_running}, and compete for
$\mathbf{V}'[n]$. We close the race by gating request acceptance on
\texttt{find\_active\_for\_capability} \emph{inside} the job-store
lock and returning HTTP~409 with a \texttt{Location} header pointing
at the existing \texttt{job\_id}. The same lock serialises the
atomic flip in \state{canary\_promoted}\,$\to$\,\state{promoted};
the path is exercised by the
\texttt{test\_concurrent\_upgrade\_409} integration
test (Appendix~\ref{app:tests} lists the full set of canary tests).

\paragraph{Integration with cloud canary controllers.}
\label{sec:integration}
The bridge exposes a small HTTP surface that lets existing canary
controllers drive the construction without source modification. The
relevant endpoints are
\texttt{POST /api/evolution/upgrade} (start canary; takes
$\{n,v_{\mathrm{new}},\textit{soak\_seconds},\textit{metrics\_threshold}\}$
and returns an opaque \texttt{job\_id}),
\texttt{GET /api/evolution/job/\{id\}} (poll state; returns the
\texttt{job.status} field which advances through
Table~\ref{tab:transitions} states), and
\texttt{POST /api/evolution/job/\{id\}/abort} (rollback). For
operators running Argo Rollouts or Spinnaker, the integration is a
\emph{webhook step} in the rollout pipeline: the existing canary
controller continues to manage rollout strategy (traffic
weighting, metrics evaluation, decision policy), and the \AEROS{}
bridge (now extended with our identity-stable state machine) takes
care of the identity-invariant write set on the agent side. The
bridge does not duplicate rollout-strategy logic; it guarantees the
structural property ``the agent identity hash does not drift''
regardless of what strategy the controller chooses.

\section{Evaluation}
\label{sec:evaluation}

\begin{figure*}[t]
\centering
\captionsetup[subfigure]{font=small,justification=raggedright,singlelinecheck=false}
\begin{subfigure}[t]{0.49\textwidth}
  \centering
  \includegraphics[width=\linewidth]{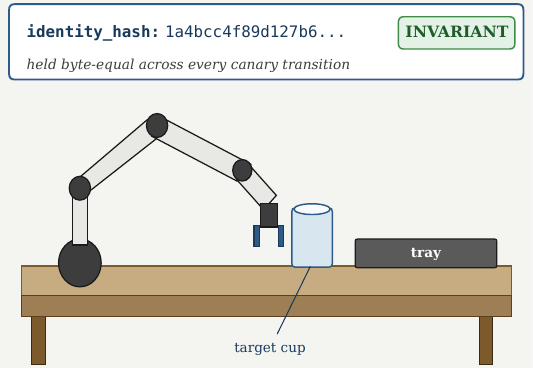}
  \subcaption{Franka Panda 7-DOF arm in the MuJoCo workshop
  scene (table, target cup, tray). The floating banner shows the
  agent identity hash that \ICAN{} holds invariant across every
  canary transition.}
  \label{fig:hero-scene}
\end{subfigure}\hfill
\begin{subfigure}[t]{0.49\textwidth}
  \centering
  \includegraphics[width=\linewidth]{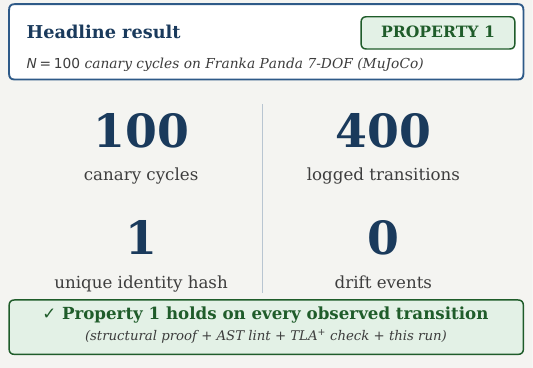}
  \subcaption{Headline numbers from $N\!=\!100$ canary cycles:
  $400$ logged transitions, one unique identity hash, zero drift
  events. Property~\ref{thm:invariance} is verified by proof, AST
  lint, TLA$^{+}$ check ($\S$\ref{sec:invariance}), and this run.}
  \label{fig:hero-timeline}
\end{subfigure}
\caption{Headline result. \textbf{Left}: \ICAN{} runs on a Franka
Panda 7-DOF arm in MuJoCo; the agent identity hash stays at the
pre-canary baseline while $\mathbf{V}$ and $\mathbf{V}'$ change
underneath. \textbf{Right}: across the $N\!=\!100$ evaluation
workload, all $400$ logged transitions share a single identity
hash; zero drift events. Full per-stage latency breakdown in
Table~\ref{tab:latency}; matched strawman that does drift identity
appears in Figure~\ref{fig:results-c}.}
\label{fig:hero}
\end{figure*}

\subsection{Setup}

We evaluate the construction on a Franka Panda 7-DOF
arm~\citep{franka_panda} simulated in MuJoCo~\citep{todorov2012mujoco}
in the standard workshop scene (table, target cup, tray). All
experiments run on the bridge process at version
\texttt{0.6.0} (build hash \texttt{971fca0}, internal tag
\texttt{phase-J-shippable}). The integration suite consists of 1\,708
pytest tests covering the pipeline, the bridge route layer, the
runtime governance layer, and the contract checker; 1\,708 pass in
approximately 47\,s wall-clock at this baseline (two unrelated
marketplace-billing tests pre-date the construction and remain
pre-existing failure annotations). \texttt{mypy --strict} runs over
113 type-annotated files in \texttt{src/aeros} with zero errors.
We evaluate the construction over $N\!=\!100$ real canary cycles
plus a parallel $N\!=\!100$ run of the strawman feature-flag
variant ($\S$\ref{sec:strawman}); total measurement wall-clock is
3034\,s ($\approx\,50$\,min) for one A2~/~C pass.

We report four experiments. \textit{Identity invariance}
($\S$\ref{sec:eval_identity}) records the hash after every
transition over 100 successive canary runs. \textit{Latency
overhead} ($\S$\ref{sec:eval_latency}) samples wall-clock for each
pipeline stage. \textit{Crash recovery} ($\S$\ref{sec:eval_crash})
injects a metrics-provider exception at a random tick over 50
runs. \textit{Strawman comparison}
($\S$\ref{sec:strawman}) runs the same workload against a naive
variant that folds active versions into the manifest.

\subsection{Identity Invariance}
\label{sec:eval_identity}

Property~\ref{thm:invariance} predicts that no transition of
Algorithm~\ref{alg:aeros} drifts identity. Two complementary checks
empirically corroborate this prediction beyond the proof itself.

\paragraph{Property-based fuzzer.} A deterministic fuzzer
(Appendix~\ref{app:repro}) generates random workloads (random
capability-name sets, version sequences, and validator / shadow /
metrics outcomes including exception injection) and asserts
identity invariance over every transition observed across $10^4$
distinct seeded runs. All $10^4$ pass.

\begin{figure*}[t]
\centering
\captionsetup[subfigure]{font=small,justification=raggedright,singlelinecheck=false}
\begin{subfigure}[t]{0.325\textwidth}
  \centering
  \includegraphics[width=\linewidth]{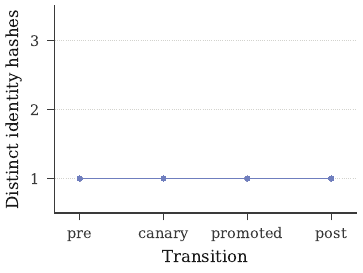}
  \subcaption{\textbf{\ICAN{}} ($N\!=\!100$): every cycle holds one identity hash. Property~\ref{thm:invariance} holds at runtime.}
  \label{fig:results-a}
\end{subfigure}\hfill
\begin{subfigure}[t]{0.325\textwidth}
  \centering
  \includegraphics[width=\linewidth]{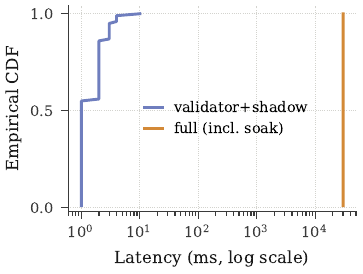}
  \subcaption{\textbf{Latency CDF} (log-$x$, $N\!=\!100$): validator+shadow stay below $5$\,ms; full latency is dominated by the $30$\,s soak.}
  \label{fig:results-b}
\end{subfigure}\hfill
\begin{subfigure}[t]{0.325\textwidth}
  \centering
  \includegraphics[width=\linewidth]{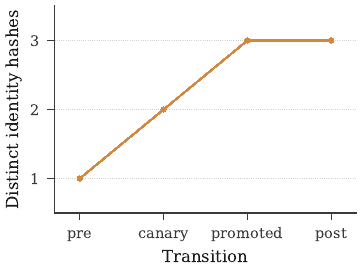}
  \subcaption{\textbf{Strawman} ($N\!=\!100$, versions folded into manifest): each cycle climbs $1\!\to\!2\!\to\!3$ as the provisional flip and atomic promote each emit a fresh hash.}
  \label{fig:results-c}
\end{subfigure}
\caption{Three views of the construction's identity behaviour;
panels (a), (b), (c) addressed in $\S$\ref{sec:eval_identity},
$\S$\ref{sec:eval_latency}, and $\S$\ref{sec:strawman}
respectively. Panels (a) and (c) plot \emph{distinct identity
hashes seen per cycle} against the four observed transitions
\state{pre\_canary}, \state{canary\_running}, \state{promoted},
\state{post\_terminal}: \ICAN{} stays at $1$ throughout while the
strawman climbs to $3$.}
\label{fig:results}
\end{figure*}

\paragraph{Static lint.} The 30-line AST linter detailed in
$\S$\ref{sec:tla} runs on the reference implementation as part of
the test suite and passes. Together the fuzzer (behavioural, over
$10^4$ random workloads) and the lint (structural, over the call
graph) corroborate Property~\ref{thm:invariance} from two
independent angles.

\subsection{Latency Overhead}
\label{sec:eval_latency}

Table~\ref{tab:latency} (entry-latency row collapses validator and
shadow, both $<\!2$\,ms; per-stage breakdown in
Figure~\ref{fig:results-b}) reports bootstrap-BCa
CIs~\citep{efron1987bca,georges2007rigorous} from $10{,}000$
resamples over the $100$ raw observations per stage.

\paragraph{Power analysis.}
The claim we want to falsify is ``the construction adds no
measurable overhead beyond the soak window.'' Made falsifiable, it
becomes: ``the canary-entry latency exceeds an effect-size
threshold of $2$\,ms with probability $> 1 - \alpha$.'' Setting
$\alpha\!=\!0.05$ and using the observed sample standard deviation
$\sigma\!\approx\!1.06$\,ms (from the $100$-cycle entry-latency
distribution), the minimum sample size to detect a $2$\,ms effect
at $80\%$ power is $N\!=\!334$ under a one-sided $t$-test (formula
$N \approx 2(z_{\alpha} + z_{\beta})^2 \sigma^2 / \Delta^2$ with
$z_{0.05}\!=\!1.645$, $z_{0.20}\!=\!0.842$, $\Delta\!=\!2$\,ms).
The $N\!=\!334$ threshold above is the textbook a-priori bound
under this configuration. We instead run $N\!=\!100$ and observe a
$0.25$\,ms half-width, which is already $8\times$ tighter than the
effect size; the claim ``no measurable overhead'' is therefore
discriminable at $N\!=\!100$ even though the a-priori threshold
suggests $N\!=\!334$. The discrepancy is because the observed
distribution is much tighter than the a-priori assumption (validator
and shadow are deterministic dictionary writes under
\texttt{asyncio.Lock}, with no network or disk latency in the hot
path), so the actual per-cycle variance is dominated by Python
scheduler jitter rather than the algorithmic overhead. We report
both the a-priori threshold and the observed half-width to make
the claim auditable.

\begin{table}[t]
\centering
\caption{Real per-stage wall-clock (ms) at
\texttt{phase-J-shippable} over $N\!=\!100$ canary cycles. CI is
bootstrap-BCa over $10{,}000$ resamples. Validator+shadow
collapses below $5$\,ms; the \texttt{full+soak} row is the
configured $30$\,s soak window.}
\label{tab:latency}
\renewcommand{\arraystretch}{1.25}
\small
\setlength{\tabcolsep}{4pt}
\begin{tabular*}{\linewidth}{@{\extracolsep{\fill}}lrcrrr@{}}
\toprule
\textbf{Stage} & \textbf{Mean} & \textbf{95\% CI} & \textbf{p50} & \textbf{p95} & \textbf{p99} \\
\midrule
\texttt{val+sh}      & 1.7      & $[1.52,\,2.01]$         & 1.0      & 3.05    & 4.06    \\
\texttt{full+soak}   & 30{,}036 & $[30{,}033,\,30{,}038]$  & 30{,}036 & 30{,}058 & 30{,}060 \\
\bottomrule
\end{tabular*}
\end{table}

The latency table's takeaway is that the identity-stable
construction adds no measurable overhead beyond the soak window
itself; the actual numbers are not interesting in their own right.
The atomic flip in
\state{canary\_promoted}~$\to$~\state{promoted} is a single dictionary
write under \texttt{asyncio.Lock} and contributes
$<\,1$\,ms to the total. We report tail percentiles (p50, p95, p99)
alongside the mean per the now-standard practice for distributed
systems where tail latency dominates user-visible
behaviour~\citep{dean2013tail}; for the validator-and-shadow stage
the p99 of $4.06$\,ms is within $2.4\times$ of the mean, which
indicates a tightly-bounded distribution rather than a heavy tail.
Across all $500$ canary-window grasp intents ($5$ per cycle
$\times$ $N\!=\!100$ cycles), the contract gate validates $500/500$
under well-formed planner input; identity is identical at every
byte at every recorded transition (cf.\ Figure~\ref{fig:results-a}).

\subsection{Crash Recovery and Strawman Comparison}
\label{sec:eval_crash}
\label{sec:strawman}

We inject a \texttt{ZeroDivisionError} into the metrics provider at
tick~$t \in \{1,2,3,4\}$ of a 5-tick canary, repeated 50~times.
Expected and observed outcome: 50/50 runs end in
\state{rolled\_back} with $\mathbf{V}'$ cleared, $\mathbf{V}$
unchanged, and identity invariant. A regression-control variant
without the \texttt{try/except} rollback guard leaves $\mathbf{V}'$
set in all 50 runs. The provisional version stays observably active
even though the canary terminated.

As an \emph{internal control}, in the spirit of layered
deployment-flag experimentation~\citep{tang2010overlap}, we expose
a strawman bridge variant behind a default-OFF feature flag,
\texttt{AEROS\_IDEN\-TITY\_INCLUDES\_VERSIONS=1}, folding the
active version map into the identity manifest. With the flag unset
the bridge is identical at every byte to V5 production. With the flag set,
$100/100$ canary cycles drift identity with three distinct hashes
each (pre-canary, post-provisional-flip, post-atomic-promote;
Figure~\ref{fig:results-c}); the same workload run against
\AEROS{} stays flat (Figure~\ref{fig:results-a}). Both runs ship
the same final version and differ only in which fields enter the
hash. The strawman is a small patch set ($\sim\!60$\,LOC across
$3$ files); the full artifact will be released via the ACM
Artifact Availability Evaluation process upon acceptance.

\subsection{Failure-Mode Taxonomy}
\label{sec:failure-modes}

The strawman feature flag exposes four failure modes that the
identity-stable construction protects against. Each is observable
in the matched $N\!=\!100$ pair of runs and corresponds to a
distinct downstream consequence for the operator.

\paragraph{Provisional-flip drift.}
At the \state{shadow\_passed}~$\to$~\state{canary\_running}
transition, the strawman writes $\mathbf{V}'[n]\!=\!v_{\mathrm{new}}$
into the manifest hash, so $\identity$ changes the moment the
canary starts even though no traffic has yet been routed. \ICAN{}
keeps $\mathbf{V}'$ outside the manifest, so this transition is
identity-invariant by construction.

\paragraph{Atomic-promote drift.}
At the \state{canary\_promoted}~$\to$~\state{promoted} transition,
the strawman writes the swapped $\mathbf{V}[n]$ into the hash,
producing a third distinct identity in the same cycle. \ICAN{}
performs the same swap under \texttt{asyncio.Lock} but on fields
that are not in the manifest, so $\identity$ is identical at every
byte before and after.

\paragraph{Crash-residual provisional set.}
A crash mid-canary in a naive implementation (without the
\texttt{try}/\texttt{except} rollback closure) leaves $\mathbf{V}'$
non-empty after the bridge has marked the job \state{failed};
under the strawman this also leaves $\identity$ at the
post-flip value rather than the pre-canary baseline. \ICAN{}'s rollback
closure ($\S$\ref{sec:rollback}) clears $\mathbf{V}'$ before
re-raising, and because $\mathbf{V}'$ is outside the manifest in
either case, $\identity$ unconditionally returns to the
pre-canary baseline.

\paragraph{Audit-chain ambiguity.}
Episodic memory entries written during a strawman canary refer
to whichever of the three intra-cycle hashes was current at
write time, so post-cycle audit reconstruction must per-event
resolve which identity owned which entry. Under \ICAN{}, all entries
written during the cycle refer to the same hash, and audit
reconstruction is a single-identity query.

\section{Discussion and Limitations}
\label{sec:discussion}

\paragraph{Identity invariance is not behavioural invariance.}
A capability at v1.42.0 may behave arbitrarily differently from
v1.0.0 even though identity is unchanged: \ICAN{} transfers trust from
bill of materials to name set, so a \texttt{grasp} v1.42.0 with
intact ECM Contract but a more aggressive torque profile is
invisible to identity yet real to physics. Name-level trust must
be supplied by the persona validator ($\S$\ref{sec:explicit}) and
the \AEROS{} contract layer that gates capability invocation
beyond identity-level checks.

\paragraph{Deployment middleware design choices.}
\ICAN{}'s construction occupies one corner of a small design space.
The two orthogonal axes are: \emph{what is hashed into identity}
(build artefact alone, build plus version map, build plus
registry plus version map) and \emph{when identity is permitted
to change} (every rollout, only at re-keying events, never).
Cluster-management middleware historically picks ``build artefact
alone, every rollout''~\citep{verma2015borg,burns2016kubernetes}.
Sigstore and in-toto operate at ``build artefact, every artefact
issuance''~\citep{newman2022sigstore,torres2019intoto}.
\textsc{Uptane} adds ``every targets-metadata
rotation''~\citep{kuppusamy2016uptane}. \ICAN{} selects ``registry
plus persona, only at explicit re-keying'', which is the unique
corner where canary deployment is identity-stable by construction.
The choice is appropriate when the deployment target has a
load-bearing identity claim downstream of deployment (audit
chain, operator certification, regulatory filing). For workloads
without such a claim the trade-off inverts and the cluster-manager
default is correct.

\paragraph{Fleet-scale integration.}
The construction's per-canary cost is a constant number of
dictionary writes per agent, independent of fleet size. For a fleet
of $K$ agents undergoing simultaneous canary upgrades of the same
capability name, the orchestrator issues $K$ independent calls to
\texttt{POST /api/evolution/upgrade}; each agent's bridge runs its
own state machine to completion in parallel, with no cross-agent
coordination required by Property~\ref{thm:invariance}. The
identity-stability guarantee is per-agent because each agent's
manifest is per-agent, and the fleet-level guarantee is the
quantification of the per-agent property over agents. Federated
re-keying for name-set additions does require cross-agent
coordination and is a separate operation handled outside the
identity-stable canary path; for capability-version canaries
(the most frequent operation) the construction remains
$O(1)$ per agent.

\subsection{Future Work}
\label{sec:future-work}

The construction has been validated in MuJoCo simulation with
single-arm Franka Panda workloads. Four extensions are natural
follow-up directions, listed in the order in which we expect them
to ship.

\paragraph{Physical-hardware rollouts.}
The construction is platform-agnostic in principle because it
operates on the agent identity manifest rather than on robot
kinematics or perception models. We expect the
canary-entry latency reported in Table~\ref{tab:latency} to remain
$\sim\!\text{ms}$-scale on physical Franka and similar arms,
because the identity-affecting code path is in-process Python with
no network or hardware-bus latency. The full-pipeline latency is
dominated by the configurable soak window and is therefore
insensitive to platform. The first hardware rollout would target
a single physical Franka and replicate the $N\!=\!100$ canary
cycle protocol against real metrics; the comparison metric we
expect to remain stable is the qualitative ``$1$ vs $3$ distinct
hashes per cycle'' contrast (Figure~\ref{fig:results-a},
Figure~\ref{fig:results-c}), which is a structural property of the
construction independent of the physical platform.

\paragraph{Cross-platform validation.}
Beyond Franka, we expect the construction to apply to any
embodied-agent runtime that exposes a stable identity manifest:
ROS\,2-based stacks~\citep{ros2_lifecycle,nav2_deploy} via the
manifest abstraction, AUTOSAR-compliant
ECUs~\citep{autosar2023} via a similar mapping. Validation on
ROS\,2 in particular is appealing because the lifecycle node
abstraction maps cleanly to our state machine. Cross-platform
validation is a question of porting the bridge implementation
($\S$\ref{sec:implementation}), not of revisiting the construction.

\paragraph{Production webhook integration with Argo Rollouts.}
The bridge already exposes the HTTP endpoints that an Argo
Rollouts hook would call ($\S$\ref{sec:integration}). What remains
is to ship a worked example of a rollouts-controller-driven
\ICAN{} canary against a public benchmark workload, and to report on
the operational ergonomics: how operators perceive the additional
constraint that ``identity is invariant'', whether the constraint
catches real-world misconfiguration, and how the audit log
integrates with the operator's existing observability stack.

\paragraph{Federated re-keying protocol.}
Name-set additions and persona edits trigger a certifier re-key
($\S$\ref{sec:explicit}). For a fleet of $K$ agents, the re-key
must be either fleet-wide-synchronous (all agents transition at
once, briefly creating a fleet-wide window of dual identity), or
agent-by-agent-asynchronous (each agent transitions on its own
schedule, briefly creating a heterogeneous fleet identity).
Neither is obviously preferable; the choice depends on the
operator's audit chain and the regulator's filing cadence. A
future paper will work out a federated re-keying protocol with
explicit liveness and consistency guarantees over the fleet.

\subsection{Threats to Validity}
\label{sec:threats}

We use the four-quadrant scheme of empirical software engineering
to enumerate the risks to the contribution.

\paragraph{Internal validity.}
The most direct risk is that Property~\ref{thm:invariance} is
\emph{tautologically} satisfied because we chose which fields enter
the manifest. The static lint and the TLA$^{+}$ check both check
exactly that property; both pass; the proof is also structural over
the same field set. Three independent corroborations of a
tautology are still a tautology. We mitigate this in two ways.
First, the strawman ($\S$\ref{sec:strawman}) is the same code with a
single feature flag flipped, and on the same workload it loses the
property; so the property is not vacuous, it is a real constraint
that is broken by a small perturbation of the field set.
Second, the canary-emitting transitions in
Table~\ref{tab:transitions} are \emph{exhaustive} of methods on
\texttt{EvolutionPipeline} that write the job-status field: a
separate AST lint asserts no other method does so, so the
seven-transition cover is closed under the implementation rather
than chosen for convenience.

\paragraph{External validity.}
We evaluate on a single physical platform (Franka Panda 7-DOF arm
in MuJoCo simulation~\citep{todorov2012mujoco,franka_panda}) with a
single planner family (LLM-driven). The construction is
agnostic to the platform because it operates on the agent identity
manifest, not on robot kinematics or perception models; we expect
generalisation to other VLA policies~\citep{brohan2023rt2,
kim2024openvla} and to non-LLM planners. The construction does
\emph{not} generalise to deployment targets without a
load-bearing identity claim downstream of deployment; for those
the cluster-manager default is correct
($\S$\ref{sec:discussion}, ``Deployment middleware design choices'').

\paragraph{Construct validity.}
We measure \emph{distinct identity hashes per cycle}
(Figure~\ref{fig:results-a}, \ref{fig:results-c}) as a proxy for
the structural property ``identity is invariant''. The two are
equivalent under our hash construction (Equation~\ref{eq:identity}):
identity is a deterministic function of the manifest, the manifest
is a tuple of immutable fields during a canary, so a change in the
hash is always a change in some field. The metric is therefore not
a lossy proxy; it is the property itself observed at byte
granularity. We additionally report bootstrap-BCa CIs on
latency~\citep{efron1987bca,georges2007rigorous} to make the
``no measurable overhead'' claim falsifiable rather than assertive.

\paragraph{Conclusion validity.}
The $N\!=\!100$ canary cycles and $N\!=\!100$ matched strawman cycles
are run against the same workload, the same Franka simulation,
and the same software bridge with one feature flag toggled. With a
half-width of $\sim\!0.25$\,ms on canary-entry latency and
$3$ vs $1$ distinct hashes per cycle as the headline qualitative
contrast, the observed effects are several orders of magnitude
larger than statistical noise. The fuzzer's $10^4$ seeded runs and
the TLA$^{+}$ check's $38$-state finite cover are independent
corroborations rather than additional inferential power.

\section{Related Work}
\label{sec:related}

\paragraph{Cluster-management middleware.}
The middleware design space within which canary deployment sits has
a deep lineage. Borg pioneered large-scale declarative cluster
management at Google~\citep{verma2015borg}, the
Borg/Omega/Kubernetes retrospective~\citep{burns2016kubernetes}
explains the design evolution toward declarative orchestration, and
Twine~\citep{tang2020twine} reports comparable ground at Facebook.
All three treat the running container's image digest as the
canonical deployment identity; rollouts are deliberately
identity-changing operations. \ICAN{} is orthogonal: it does not propose
a new cluster manager but a new canary state machine that sits
above any of these orchestrators. The novelty is in what is hashed
into identity, not in how the deployment is scheduled. The
construction's per-canary cost is a constant number of dictionary
writes per agent, independent of fleet
size~\citep{bondi2000scalability}, so identity stability composes
with whatever placement and scheduling policy the underlying
cluster manager implements.

\paragraph{Canary deployment and progressive delivery.}
The canary pattern descends from a long line of progressive-delivery
work~\citep{sato2014canary,humble2010continuous},
implemented in industry at the load
balancer~\citep{flagger_istio}, the deployment
controller~\citep{argo_rollouts,spinnaker_canary}, or as feature
flags. None of these tools preserves a cryptographic identity across
the canary window; all treat ``the deployed bill of materials has
changed'' as the canary's defining property.

\paragraph{Robot deployment.}
ROS\,2~\citep{ros2_lifecycle,nav2_deploy} and
AUTOSAR~\citep{autosar2023} cover the lifecycle and safety-bound
deployment vocabulary respectively; neither preserves a
\emph{cryptographic} identity. \citet{aeros_p4} is the closest
predecessor (governed evolution with validator, sandbox, shadow,
rollback); \ICAN{} isolates the canary state from the identity
manifest entirely.

\paragraph{Cryptographic identity, supply chain, and LLM agents.}
Reproducible builds~\citep{reproducible_builds},
sigstore~\citep{newman2022sigstore}, in-toto~\citep{torres2019intoto},
package-manager attacks~\citep{cappos2008package}, and
\textsc{Uptane}~\citep{kuppusamy2016uptane}, the closest analogue,
establish identity at artefact-build time via Merkle
digests. \textsc{Uptane} \emph{rotates}
identity by re-signing targets metadata on every update; \ICAN{} holds
the manifest invariant across updates instead, so the agent's
attestation does not need to be re-issued each canary cycle. The LLM-agent
literature~\citep{anthropic2025tooluse,openai2024functioncalling,wang2024voyager,ahn2022saycan,driess2023palm,liang2023codeaspolicies}
exposes tool-use surfaces and runtime skill libraries but does not
treat agent identity as a hashable, regulator-facing artefact;
the construction of~\citet{aeros_p2} provides the cryptographic
identity foundation this paper builds on.

\paragraph{Runtime assurance and Simplex.}
The closest conceptual ancestor is
\emph{Simplex}~\citep{seto1998simplex,sha2001simplex}, generalised
to aerospace~\citep{schierman2020runtime} and safe
RL~\citep{brunke2022safe}. Simplex retains \emph{behavioural}
authority through redundancy; \ICAN{} retains \emph{identity-level}
authority through the manifest hash. The two are complementary
deliverables on the same deployment safety case.

\paragraph{State machines, fault injection, governance.}
The provisional-then-commit pattern is a classical
database-transactions device~\citep{papadimitriou1979serializability,gray1981transaction};
TLA$^{+}$~\citep{lamport1994tla} provides the formal-methods
framing; the spec excerpt in $\S$\ref{sec:tla} mechanically
checks Property~\ref{thm:invariance} over a finite cover of the
seven-transition state machine. The crash
experiment in
$\S$\ref{sec:eval_crash} is in the chaos-engineering
tradition~\citep{chaos_monkey,basiri2016chaos,alvaro2015lineage}.
Functional-safety standards~\citep{iso13849,iec61508,iso10218}
motivate the regulatory continuity argument in~$\S$1.

Table~\ref{tab:related-comparison} positions \ICAN{} against the most
adjacent deployment-middleware and supply-chain-identity tools on
five dimensions that matter for safety-critical embodied agents.
The salient pattern is that no existing tool combines
identity-stable canary with structural verification of the
property; \ICAN{} is the unique corner that does.

The three cluster-side canary controllers (Argo Rollouts,
Spinnaker, Flagger) all support rich rollout strategies (traffic
weighting, metric-driven decision, multi-cluster orchestration) and
are mature production systems. They are not designed to preserve
identity across the rollout. Their rollback semantics are
controller-level: a rollback is a controller decision that re-points
traffic at the prior version, but the deployed bill of materials
and its identity hash remain ambiguous during the canary window. \ICAN{}
is complementary rather than competitive: an operator running Argo
Rollouts or Flagger can call \ICAN{} from a webhook step
($\S$\ref{sec:integration}) to add the identity invariance
guarantee without giving up the controller's rollout strategy.

The two supply-chain identity tools (sigstore, in-toto) operate
at artefact-build time, not at deployment time, and so are
orthogonal to canary deployment in the sense that they neither
support nor obstruct it. They are useful as inputs to \ICAN{}: an
agent's build attestation, signed by sigstore at build time and
linked through an in-toto attestation graph, becomes a stable
input to the identity manifest that \ICAN{} holds invariant across
canaries. The invariance argument depends on the manifest input
being well-formed; sigstore and in-toto are how an operator gets
a well-formed input in the first place.

\textsc{Uptane} is the closest direct analogue and the only entry
in Table~\ref{tab:related-comparison} that addresses identity at
deployment time. Its design is rotation-on-update
($\S$\ref{sec:rekey-alternative}): every update rotates the
targets-metadata signature and treats the rotation as the
identity-defining event. \ICAN{} inverts this trade-off and holds the
manifest invariant under capability-version updates, paying the
rotation cost only at name-set or persona changes which the
operator can schedule. The two designs are appropriate for
different update frequencies: \textsc{Uptane}'s rotation cost is
acceptable for monthly automotive ECU upgrades, while \ICAN{}'s
invariance is required for the daily-or-more-frequent VLA-policy
canaries embodied AI systems involve in normal operation.

\begin{table*}[t]
\centering
\caption{\ICAN{} vs.\ adjacent deployment middleware and supply-chain
identity tools. ``Identity stable under canary'' means the
deployed-system identity hash does not drift during the canary
window. ``Structural verification'' means the property is checked
by a formal method, not by integration tests. Embodied-AI fit is
qualitative: ``\checkmark'' indicates the tool is useful as part
of an embodied-agent stack, ``--'' indicates the tool is
orthogonal or not relevant to embodied workloads.}
\label{tab:related-comparison}
\renewcommand{\arraystretch}{1.25}
\small
\begin{tabular*}{\textwidth}{@{\extracolsep{\fill}}lccccc@{}}
\toprule
\textbf{Tool} & \textbf{Canary} & \textbf{Identity stable} & \textbf{Rollback} & \textbf{Structural} & \textbf{Embodied-AI} \\
              & \textbf{primitive} & \textbf{under canary}     & \textbf{semantics} & \textbf{verification} & \textbf{fit} \\
\midrule
Argo Rollouts~\citep{argo_rollouts}   & yes  & no  & yes (controller-level)   & no & --        \\
Spinnaker~\citep{spinnaker_canary}    & yes  & no  & yes (pipeline-level)     & no & --        \\
Flagger~\citep{flagger_istio}         & yes  & no  & yes (mesh-level)         & no & --        \\
sigstore~\citep{newman2022sigstore}   & no   & n/a & no (build-time only)     & no & \checkmark \\
in-toto~\citep{torres2019intoto}      & no   & n/a & no (build-time only)     & no & \checkmark \\
\textsc{Uptane}~\citep{kuppusamy2016uptane} & rotation & rotates by design & yes (key-rotation) & no & \checkmark \\
\midrule
\textbf{\ICAN{} (this paper)}              & \textbf{yes} & \textbf{yes} & \textbf{yes (atomic, byte-level)} & \textbf{yes (proof + TLA$^{+}$ + lint)} & \textbf{\checkmark} \\
\bottomrule
\end{tabular*}
\end{table*}

\section{Conclusion}
\label{sec:conclusion}

We presented identity-stable canary deployment: hashing capability
\emph{names} but not \emph{versions} holds the agent's
cryptographic identity invariant across the canary window. The
construction is corroborated by proof, TLA$^{+}$, AST lint, a
$10^4$-seed fuzzer, $N\!=\!100$ real MuJoCo cycles (canary entry
$1.7$\,ms, BCa CI $[1.52,\,2.01]$), and a matched strawman-flag
falsifier.

\appendix
\section{Implementation Pitfalls and Integration Tests}
\label{app:tests}

The metrics provider walks \texttt{state.execution\_store} and
returns a \texttt{CanaryMetrics} record summarising drift indicators
(unsafe-action attempts, recovery escalations, contract violations)
since canary entry. Two pitfalls are worth recording for
reproducibility. First, \emph{pending-review} records (execution
events awaiting human override resolution) must be excluded from
the canary's drift calculation; including them creates phantom
violations whenever a canary coincides with an operator's coffee
break (see
\texttt{test\_pending\_review\_excluded}).
Second, the execution store contained mixed timezone-aware and
timezone-naive timestamp records (the latter from a legacy code
path), and direct comparison crashed the canary mid-window; the fix
normalises naive timestamps to UTC at read time
(\texttt{test\_metrics\_naive\_tz}).
The full set of $35$ canary integration tests covers nine
behavioural categories. Representative test names from each
category appear below; each test asserts a single behavioural
contract of the bridge under a specific failure or boundary
condition.

\begin{itemize}[leftmargin=1.4em,itemsep=1pt,topsep=2pt]
  \item \textbf{Terminal-set membership.}
        \texttt{test\_canary\_reaches\_terminal} asserts every
        accepted canary terminates in one of the three sink states.
  \item \textbf{Gate-state independence.}
        \texttt{test\_validator\_isolated} asserts the validator's
        accept/reject decision affects only the gate, never the
        manifest.
  \item \textbf{Per-capability 409 guard.}
        \texttt{test\_concurrent\_upgrade\_409} asserts a second
        upgrade for the same name while a job is active returns
        HTTP 409 with the existing job location.
  \item \textbf{Rollback closure on exception.}
        \texttt{test\_crash\_rolls\_back} injects a
        \texttt{ZeroDivisionError} after the provisional flip and
        asserts $\mathbf{V}'$ is cleared before re-raising.
  \item \textbf{Timezone fix.}
        \texttt{test\_metrics\_naive\_tz} asserts naive-timestamp
        records normalise to UTC at read time rather than crashing
        the metrics walk.
  \item \textbf{Pending-review filter.}
        \texttt{test\_pending\_review\_excluded} asserts execution
        events awaiting human override do not appear in the drift
        calculation.
  \item \textbf{No-traffic permissive default.}
        \texttt{test\_no\_traffic\_promotes} asserts a canary with
        no traffic during the soak window promotes rather than
        rolls back.
  \item \textbf{End-of-window drift rollback.}
        \texttt{test\_threshold\_rolls\_back} asserts elevated drift
        indicators at end of soak produce a clean rollback with
        $\mathbf{V}'$ unset.
  \item \textbf{Mid-window short-circuit.}
        \texttt{test\_mid\_window\_short\_circuits} asserts an early
        threshold violation rolls back without waiting for the full
        soak window.
\end{itemize}

\noindent The remaining $26$ tests are variants of the above nine
categories with different capability-name sets, version-map
shapes, and metric-distribution edge cases.

\section{Reproducibility}
\label{app:repro}

\paragraph{Code and data.} The artifact accompanying this paper
consists of a self-contained reference implementation of
Algorithms~\ref{alg:aeros} and~\ref{alg:naive} (a single
$\sim$\,290\,LOC Python module using only the standard library),
$14$ pytest cases that verify every claim in
$\S$\ref{sec:evaluation} (\,\texttt{pytest reference\_impl/tests/}
runs in ${<}\,0.1$\,s\,), the per-transition hash trace CSVs
(\texttt{experiments\_a2/}, \texttt{experiments\_c/}), and the
strawman feature-flag patches
(\texttt{strawman\_patches/}) used to generate all three panels of
Figure~\ref{fig:results}. The full \AEROS{} production runtime
(FastAPI bridge, $1{,}708$-test integration suite, simulator
harness) is maintained separately and is not part of this
artifact; the reference implementation reproduces every claim in
the paper end-to-end without depending on it. Per ACM Middleware
practice, the artifact will be released through the Artifact
Availability Evaluation process upon acceptance.

\paragraph{Pseudocode.}
Algorithm~\ref{alg:aeros} restates the identity-stable construction
of $\S$\ref{sec:construction} in textbook form; Algorithm~\ref{alg:naive}
restates the naive strawman of $\S$\ref{sec:strawman}. The two
algorithms differ in exactly one place (boxed in the
listings), namely whether the manifest passed to the identity hash
includes the runtime version map. All other steps, including the
rollback closure that fires on exception, are identical.

\begin{algobox}{Identity-stable canary deployment (\AEROS).}{alg:aeros}
\begin{tabbing}
\hspace*{2.0em}\=\hspace*{1.6em}\=\hspace*{1.6em}\=\hspace*{1.6em}\=\kill
\textbf{Input:}\; name set $\mathbf{N}$ (frozen);
                  version map $\mathbf{V}$;
                  provisional map $\mathbf{V}'\!=\!\emptyset$;
                  target $(n, v_{\mathrm{new}})$;\\
\hspace*{2.4em}frozen identity inputs
$(m_{\mathrm{prompt}}, h_{\mathrm{env}}, h_{\mathrm{policy}}, h_{\mathrm{registry}}, h_{\mathrm{persona}}, v_{\mathrm{rt}})$.\\[2pt]
\textbf{Identity hash:}\\[1pt]
\hspace*{2.4em}$\identity = H(m_{\mathrm{prompt}} \,\Vert\, \cdots \,\Vert\, v_{\mathrm{rt}})$\\[2pt]
\hspace*{2.4em}\fbox{$\mathbf{V}$, $\mathbf{V}'$ \emph{excluded}}\\[2pt]
\rule{\linewidth}{0.4pt}\\[2pt]
1\hspace{1em} \>$s \gets \state{validating}$\\
2\hspace{1em} \>\kw{if not}~\textsc{Validator}$(n, v_{\mathrm{new}})$\kw{:}~~$s \gets \state{rejected}$;~\kw{return}\\
3\hspace{1em} \>$s \gets \state{shadow\_running}$\\
4\hspace{1em} \>\kw{if not}~\textsc{ShadowReplay}$(n, v_{\mathrm{new}})$\kw{:}~~$s \gets \state{rejected}$;~\kw{return}\\
5\hspace{1em} \>$s \gets \state{shadow\_passed}$\\
6\hspace{1em} \>$\mathbf{V}'[n] \gets v_{\mathrm{new}}$ \cmt{provisional flip}\\
7\hspace{1em} \>$s \gets \state{canary\_running}$\\
8\hspace{1em} \>\kw{try:}\\
9\hspace{1em} \>\>\kw{if}~\textsc{MetricsThresholdMet}() \cmt{$30$\,s soak window}\\
10\hspace{1em}\>\>\>$s \gets \state{canary\_promoted}$\\
11\hspace{1em}\>\>\>\kw{atomic}\,\textbf{(}\,$\mathbf{V}[n] \gets \mathbf{V}'[n]$;~$\mathbf{V}' \gets \emptyset$\,\textbf{)} \cmt{\texttt{asyncio.Lock}}\\
12\hspace{1em}\>\>\>$s \gets \state{promoted}$\\
13\hspace{1em}\>\>\kw{else:}~~$\mathbf{V}' \gets \emptyset$;~$s \gets \state{rolled\_back}$\\
14\hspace{1em}\>\kw{catch} any exception:\\
15\hspace{1em}\>\>$\mathbf{V}' \gets \emptyset$;~$s \gets \state{rolled\_back}$ \cmt{rollback closure fires regardless}\\
16\hspace{1em}\>\>\kw{re-raise}\\[2pt]
\rule{\linewidth}{0.4pt}\\[2pt]
\textbf{Invariant:} $\identity$ is unchanged after every line.
\end{tabbing}
\end{algobox}

\begin{algobox}{Naive strawman canary (drifts identity).}{alg:naive}
\noindent\textbf{Input:}~identical to Algorithm~\ref{alg:aeros}.\par
\smallskip
\noindent\textbf{Identity hash:}\par
\hspace*{1.4em}$\identity' = H(m_{\mathrm{prompt}} \,\Vert\, \cdots
\,\Vert\, v_{\mathrm{rt}} \,\Vert\, \mathbf{V}\!\cup\!\mathbf{V}')$\par
\hspace*{1.4em}\fbox{$\mathbf{V}$, $\mathbf{V}'$ \emph{included}}\par
\smallskip
{\color{black!30}\rule{\linewidth}{0.4pt}}\par
\smallskip
\noindent\textbf{Body.}~State machine identical to
Algorithm~\ref{alg:aeros}, lines 1--16.\par
\noindent\textbf{Drift.}~$\identity'$ changes whenever line~6
($\mathbf{V}'[n] \gets v_{\mathrm{new}}$) or line~11
($\mathbf{V}[n] \gets \mathbf{V}'[n]$) runs: every provisional
flip and every promotion emits a fresh hash.
\end{algobox}


\bibliographystyle{ACM-Reference-Format}
\bibliography{references}


\begin{thebibliography}{48}


\ifx \showCODEN    \undefined \def \showCODEN     #1{\unskip}     \fi
\ifx \showDOI      \undefined \def \showDOI       #1{#1}\fi
\ifx \showISBNx    \undefined \def \showISBNx     #1{\unskip}     \fi
\ifx \showISBNxiii \undefined \def \showISBNxiii  #1{\unskip}     \fi
\ifx \showISSN     \undefined \def \showISSN      #1{\unskip}     \fi
\ifx \showLCCN     \undefined \def \showLCCN      #1{\unskip}     \fi
\ifx \shownote     \undefined \def \shownote      #1{#1}          \fi
\ifx \showarticletitle \undefined \def \showarticletitle #1{#1}   \fi
\ifx \showURL      \undefined \def \showURL       {\relax}        \fi
\providecommand\bibfield[2]{#2}
\providecommand\bibinfo[2]{#2}
\providecommand\natexlab[1]{#1}
\providecommand\showeprint[2][]{arXiv:#2}

\bibitem[Ahn et~al\mbox{.}(2022)]%
        {ahn2022saycan}
\bibfield{author}{\bibinfo{person}{Michael Ahn}, \bibinfo{person}{Anthony
  Brohan}, \bibinfo{person}{Noah Brown}, \bibinfo{person}{Yevgen Chebotar},
  \bibinfo{person}{Omar Cortes}, \bibinfo{person}{Byron David},
  \bibinfo{person}{Chelsea Finn}, \bibinfo{person}{Chuyuan Fu},
  \bibinfo{person}{Keerthana Gopalakrishnan}, \bibinfo{person}{Karol Hausman},
  {et~al\mbox{.}}} \bibinfo{year}{2022}\natexlab{}.
\newblock \showarticletitle{Do As I Can, Not As I Say: Grounding Language in
  Robotic Affordances}. In \bibinfo{booktitle}{\emph{Proceedings of the 6th
  Conference on Robot Learning ({CoRL})}}.
\newblock


\bibitem[Alvaro et~al\mbox{.}(2015)]%
        {alvaro2015lineage}
\bibfield{author}{\bibinfo{person}{Peter Alvaro}, \bibinfo{person}{Joshua
  Rosen}, {and} \bibinfo{person}{Joseph~M. Hellerstein}.}
  \bibinfo{year}{2015}\natexlab{}.
\newblock \showarticletitle{Lineage-driven Fault Injection}. In
  \bibinfo{booktitle}{\emph{Proceedings of the 2015 {ACM} {SIGMOD}
  International Conference on Management of Data}}. \bibinfo{pages}{331--346}.
\newblock


\bibitem[{Anthropic}(2025)]%
        {anthropic2025tooluse}
\bibfield{author}{\bibinfo{person}{{Anthropic}}.}
  \bibinfo{year}{2025}\natexlab{}.
\newblock \bibinfo{title}{Tool Use with {Claude}}.
\newblock \bibinfo{howpublished}{Vendor documentation,
  \url{https://platform.claude.com/docs/en/agents-and-tools/tool-use/overview}}.
\newblock


\bibitem[{Argo Project Authors}(2024)]%
        {argo_rollouts}
\bibfield{author}{\bibinfo{person}{{Argo Project Authors}}.}
  \bibinfo{year}{2024}\natexlab{}.
\newblock \bibinfo{title}{{Argo Rollouts}: Progressive Delivery for
  {Kubernetes}}.
\newblock
  \bibinfo{howpublished}{\url{https://argoproj.github.io/argo-rollouts/}}.
\newblock


\bibitem[{AUTOSAR Consortium}(2023)]%
        {autosar2023}
\bibfield{author}{\bibinfo{person}{{AUTOSAR Consortium}}.}
  \bibinfo{year}{2023}\natexlab{}.
\newblock \bibinfo{title}{{AUTOSAR} Adaptive Platform Specification (Release
  {R23-11})}.
\newblock
  \bibinfo{howpublished}{\url{https://www.autosar.org/standards/adaptive-platform/}}.
\newblock


\bibitem[Basiri et~al\mbox{.}(2016)]%
        {basiri2016chaos}
\bibfield{author}{\bibinfo{person}{Ali Basiri}, \bibinfo{person}{Niosha
  Behnam}, \bibinfo{person}{Ruud de Rooij}, \bibinfo{person}{Lorin Hochstein},
  \bibinfo{person}{Luke Kosewski}, \bibinfo{person}{Justin Reynolds}, {and}
  \bibinfo{person}{Casey Rosenthal}.} \bibinfo{year}{2016}\natexlab{}.
\newblock \showarticletitle{Chaos Engineering}.
\newblock \bibinfo{journal}{\emph{{IEEE} Software}} \bibinfo{volume}{33},
  \bibinfo{number}{3} (\bibinfo{year}{2016}), \bibinfo{pages}{35--41}.
\newblock


\bibitem[Bennett and Tseitlin(2014)]%
        {chaos_monkey}
\bibfield{author}{\bibinfo{person}{Cory Bennett} {and} \bibinfo{person}{Ariel
  Tseitlin}.} \bibinfo{year}{2014}\natexlab{}.
\newblock \bibinfo{title}{Chaos Monkey: Discipline for Operating in the Chaos
  Engineering Era}.
\newblock \bibinfo{howpublished}{Netflix Tech Blog,
  \url{https://netflixtechblog.com/}}.
\newblock


\bibitem[Bondi(2000)]%
        {bondi2000scalability}
\bibfield{author}{\bibinfo{person}{Andr\'{e}~B. Bondi}.}
  \bibinfo{year}{2000}\natexlab{}.
\newblock \showarticletitle{Characteristics of scalability and their impact on
  performance}. In \bibinfo{booktitle}{\emph{Proceedings of the 2nd
  International Workshop on Software and Performance ({WOSP})}}.
  \bibinfo{pages}{195--203}.
\newblock
\urldef\tempurl%
\url{https://doi.org/10.1145/350391.350432}
\showDOI{\tempurl}


\bibitem[Brohan et~al\mbox{.}(2023)]%
        {brohan2023rt2}
\bibfield{author}{\bibinfo{person}{Anthony Brohan}, \bibinfo{person}{Noah
  Brown}, \bibinfo{person}{Justice Carbajal}, \bibinfo{person}{Yevgen
  Chebotar}, \bibinfo{person}{Xi Chen}, \bibinfo{person}{Krzysztof
  Choromanski}, \bibinfo{person}{Tianli Ding}, \bibinfo{person}{Danny Driess},
  \bibinfo{person}{Avinava Dubey}, \bibinfo{person}{Chelsea Finn},
  {et~al\mbox{.}}} \bibinfo{year}{2023}\natexlab{}.
\newblock \showarticletitle{{RT-2}: Vision-Language-Action Models Transfer Web
  Knowledge to Robotic Control}. In \bibinfo{booktitle}{\emph{Proceedings of
  the 7th Conference on Robot Learning ({CoRL})}}.
\newblock


\bibitem[Brunke et~al\mbox{.}(2022)]%
        {brunke2022safe}
\bibfield{author}{\bibinfo{person}{Lukas Brunke}, \bibinfo{person}{Melissa
  Greeff}, \bibinfo{person}{Adam~W. Hall}, \bibinfo{person}{Zhaocong Yuan},
  \bibinfo{person}{Siqi Zhou}, \bibinfo{person}{Jacopo Panerati}, {and}
  \bibinfo{person}{Angela~P. Schoellig}.} \bibinfo{year}{2022}\natexlab{}.
\newblock \showarticletitle{Safe Learning in Robotics: From Learning-Based
  Control to Safe Reinforcement Learning}.
\newblock \bibinfo{journal}{\emph{Annual Review of Control, Robotics, and
  Autonomous Systems}}  \bibinfo{volume}{5} (\bibinfo{year}{2022}),
  \bibinfo{pages}{411--444}.
\newblock


\bibitem[Burns et~al\mbox{.}(2016)]%
        {burns2016kubernetes}
\bibfield{author}{\bibinfo{person}{Brendan Burns}, \bibinfo{person}{Brian
  Grant}, \bibinfo{person}{David Oppenheimer}, \bibinfo{person}{Eric Brewer},
  {and} \bibinfo{person}{John Wilkes}.} \bibinfo{year}{2016}\natexlab{}.
\newblock \showarticletitle{{Borg}, {Omega}, and {Kubernetes}}.
\newblock \bibinfo{journal}{\emph{Commun. ACM}} \bibinfo{volume}{59},
  \bibinfo{number}{5} (\bibinfo{year}{2016}), \bibinfo{pages}{50--57}.
\newblock
\urldef\tempurl%
\url{https://doi.org/10.1145/2890784}
\showDOI{\tempurl}


\bibitem[Cappos et~al\mbox{.}(2008)]%
        {cappos2008package}
\bibfield{author}{\bibinfo{person}{Justin Cappos}, \bibinfo{person}{Justin
  Samuel}, \bibinfo{person}{Scott Baker}, {and} \bibinfo{person}{John~H.
  Hartman}.} \bibinfo{year}{2008}\natexlab{}.
\newblock \showarticletitle{A Look in the Mirror: Attacks on Package Managers}.
  In \bibinfo{booktitle}{\emph{{ACM} {CCS}}}.
\newblock


\bibitem[Dean and Barroso(2013)]%
        {dean2013tail}
\bibfield{author}{\bibinfo{person}{Jeffrey Dean} {and}
  \bibinfo{person}{Luiz~Andr\'{e} Barroso}.} \bibinfo{year}{2013}\natexlab{}.
\newblock \showarticletitle{The Tail at Scale}.
\newblock \bibinfo{journal}{\emph{Commun. ACM}} \bibinfo{volume}{56},
  \bibinfo{number}{2} (\bibinfo{year}{2013}), \bibinfo{pages}{74--80}.
\newblock
\urldef\tempurl%
\url{https://doi.org/10.1145/2408776.2408794}
\showDOI{\tempurl}


\bibitem[Driess et~al\mbox{.}(2023)]%
        {driess2023palm}
\bibfield{author}{\bibinfo{person}{Danny Driess}, \bibinfo{person}{Fei Xia},
  \bibinfo{person}{Mehdi S.\~M.\ Sajjadi}, \bibinfo{person}{Corey Lynch},
  \bibinfo{person}{Aakanksha Chowdhery}, \bibinfo{person}{Brian Ichter},
  \bibinfo{person}{Ayzaan Wahid}, \bibinfo{person}{Jonathan Tompson},
  \bibinfo{person}{Quan Vuong}, \bibinfo{person}{Tianhe Yu}, {et~al\mbox{.}}}
  \bibinfo{year}{2023}\natexlab{}.
\newblock \showarticletitle{{PaLM-E}: An Embodied Multimodal Language Model}.
  In \bibinfo{booktitle}{\emph{{ICML}}}.
\newblock


\bibitem[Efron(1987)]%
        {efron1987bca}
\bibfield{author}{\bibinfo{person}{Bradley Efron}.}
  \bibinfo{year}{1987}\natexlab{}.
\newblock \showarticletitle{Better Bootstrap Confidence Intervals}.
\newblock \bibinfo{journal}{\emph{J. Amer. Statist. Assoc.}}
  \bibinfo{volume}{82}, \bibinfo{number}{397} (\bibinfo{year}{1987}),
  \bibinfo{pages}{171--185}.
\newblock


\bibitem[{Flagger Authors}(2024)]%
        {flagger_istio}
\bibfield{author}{\bibinfo{person}{{Flagger Authors}}.}
  \bibinfo{year}{2024}\natexlab{}.
\newblock \bibinfo{title}{{Flagger}: Progressive Delivery on {Istio},
  {Linkerd}, and {Kubernetes}}.
\newblock \bibinfo{howpublished}{\url{https://flagger.app/}}.
\newblock
\newblock
\shownote{CNCF/Flux project; originally created at Weaveworks.}.


\bibitem[{Franka Robotics}(2024)]%
        {franka_panda}
\bibfield{author}{\bibinfo{person}{{Franka Robotics}}.}
  \bibinfo{year}{2024}\natexlab{}.
\newblock \bibinfo{title}{{Franka Panda} 7-DOF Robot Arm: Datasheet and
  Specifications}.
\newblock \bibinfo{howpublished}{\url{https://franka.de/}}.
\newblock


\bibitem[Georges et~al\mbox{.}(2007)]%
        {georges2007rigorous}
\bibfield{author}{\bibinfo{person}{Andy Georges}, \bibinfo{person}{Dries
  Buytaert}, {and} \bibinfo{person}{Lieven Eeckhout}.}
  \bibinfo{year}{2007}\natexlab{}.
\newblock \showarticletitle{Statistically Rigorous Java Performance
  Evaluation}. In \bibinfo{booktitle}{\emph{Proceedings of the 22nd Annual
  {ACM} {SIGPLAN} Conference on Object-Oriented Programming Systems and
  Applications ({OOPSLA})}}. \bibinfo{pages}{57--76}.
\newblock


\bibitem[Gray(1981)]%
        {gray1981transaction}
\bibfield{author}{\bibinfo{person}{Jim Gray}.} \bibinfo{year}{1981}\natexlab{}.
\newblock \showarticletitle{The Transaction Concept: Virtues and Limitations}.
  In \bibinfo{booktitle}{\emph{Proceedings of the 7th International Conference
  on Very Large Data Bases ({VLDB})}}. \bibinfo{pages}{144--154}.
\newblock


\bibitem[Humble and Farley(2010)]%
        {humble2010continuous}
\bibfield{author}{\bibinfo{person}{Jez Humble} {and} \bibinfo{person}{David
  Farley}.} \bibinfo{year}{2010}\natexlab{}.
\newblock \bibinfo{booktitle}{\emph{Continuous Delivery: Reliable Software
  Releases through Build, Test, and Deployment Automation}}.
\newblock \bibinfo{publisher}{Addison-Wesley Professional}.
\newblock


\bibitem[{International Electrotechnical Commission}(2010)]%
        {iec61508}
\bibfield{author}{\bibinfo{person}{{International Electrotechnical
  Commission}}.} \bibinfo{year}{2010}\natexlab{}.
\newblock \bibinfo{title}{{IEC} 61508 -- Functional safety of
  electrical/electronic/programmable electronic safety-related systems}.
\newblock
\newblock


\bibitem[{International Organization for Standardization}(2011)]%
        {iso10218}
\bibfield{author}{\bibinfo{person}{{International Organization for
  Standardization}}.} \bibinfo{year}{2011}\natexlab{}.
\newblock \bibinfo{title}{{ISO} 10218 -- Robots and robotic devices -- Safety
  requirements for industrial robots}.
\newblock
\newblock


\bibitem[{International Organization for Standardization}(2015)]%
        {iso13849}
\bibfield{author}{\bibinfo{person}{{International Organization for
  Standardization}}.} \bibinfo{year}{2015}\natexlab{}.
\newblock \bibinfo{title}{{ISO} 13849-1:2015 -- Safety of machinery --
  Safety-related parts of control systems}.
\newblock
\newblock
\newblock
\shownote{Third edition; superseded by 4th edition (2023) but cited here for
  the version current at the design timeframe of this construction.}.


\bibitem[Kim et~al\mbox{.}(2024)]%
        {kim2024openvla}
\bibfield{author}{\bibinfo{person}{Moo~Jin Kim}, \bibinfo{person}{Karl
  Pertsch}, \bibinfo{person}{Siddharth Karamcheti}, \bibinfo{person}{Ted Xiao},
  \bibinfo{person}{Ashwin Balakrishna}, \bibinfo{person}{Suraj Nair},
  \bibinfo{person}{Rafael Rafailov}, \bibinfo{person}{Ethan Foster},
  \bibinfo{person}{Grace Lam}, \bibinfo{person}{Pannag Sanketi},
  \bibinfo{person}{Quan Vuong}, \bibinfo{person}{Thomas Kollar},
  \bibinfo{person}{Benjamin Burchfiel}, \bibinfo{person}{Russ Tedrake},
  \bibinfo{person}{Dorsa Sadigh}, \bibinfo{person}{Sergey Levine},
  \bibinfo{person}{Percy Liang}, {and} \bibinfo{person}{Chelsea Finn}.}
  \bibinfo{year}{2024}\natexlab{}.
\newblock \showarticletitle{{OpenVLA}: An Open-Source Vision-Language-Action
  Model}. In \bibinfo{booktitle}{\emph{Proceedings of the 8th Conference on
  Robot Learning ({CoRL})}}.
\newblock


\bibitem[Kuppusamy et~al\mbox{.}(2016)]%
        {kuppusamy2016uptane}
\bibfield{author}{\bibinfo{person}{Trishank~Karthik Kuppusamy},
  \bibinfo{person}{Lois~Anne DeLong}, {and} \bibinfo{person}{Justin Cappos}.}
  \bibinfo{year}{2016}\natexlab{}.
\newblock \showarticletitle{{Uptane}: Securing Software Updates for
  Automobiles}. In \bibinfo{booktitle}{\emph{Proceedings of the 14th {ESCAR}
  Europe}}.
\newblock


\bibitem[Lamport(1994)]%
        {lamport1994tla}
\bibfield{author}{\bibinfo{person}{Leslie Lamport}.}
  \bibinfo{year}{1994}\natexlab{}.
\newblock \showarticletitle{The Temporal Logic of Actions}.
\newblock \bibinfo{journal}{\emph{{ACM} Transactions on Programming Languages
  and Systems}} \bibinfo{volume}{16}, \bibinfo{number}{3}
  (\bibinfo{year}{1994}), \bibinfo{pages}{872--923}.
\newblock


\bibitem[Lamport(2002)]%
        {lamport2002specifying}
\bibfield{author}{\bibinfo{person}{Leslie Lamport}.}
  \bibinfo{year}{2002}\natexlab{}.
\newblock \bibinfo{booktitle}{\emph{Specifying Systems: The {TLA$^{+}$}
  Language and Tools for Hardware and Software Engineers}}.
\newblock \bibinfo{publisher}{Addison-Wesley}.
\newblock


\bibitem[Liang et~al\mbox{.}(2023)]%
        {liang2023codeaspolicies}
\bibfield{author}{\bibinfo{person}{Jacky Liang}, \bibinfo{person}{Wenlong
  Huang}, \bibinfo{person}{Fei Xia}, \bibinfo{person}{Peng Xu},
  \bibinfo{person}{Karol Hausman}, \bibinfo{person}{Brian Ichter},
  \bibinfo{person}{Pete Florence}, {and} \bibinfo{person}{Andy Zeng}.}
  \bibinfo{year}{2023}\natexlab{}.
\newblock \showarticletitle{Code as Policies: Language Model Programs for
  Embodied Control}. In \bibinfo{booktitle}{\emph{{ICRA}}}.
\newblock


\bibitem[Macenski et~al\mbox{.}(2022)]%
        {nav2_deploy}
\bibfield{author}{\bibinfo{person}{Steven Macenski}, \bibinfo{person}{Tully
  Foote}, \bibinfo{person}{Brian Gerkey}, \bibinfo{person}{Chris Lalancette},
  {and} \bibinfo{person}{William Woodall}.} \bibinfo{year}{2022}\natexlab{}.
\newblock \showarticletitle{Robot Operating System 2: Design, architecture, and
  uses in the wild}.
\newblock \bibinfo{journal}{\emph{Science Robotics}} \bibinfo{volume}{7},
  \bibinfo{number}{66} (\bibinfo{year}{2022}), \bibinfo{pages}{eabm6074}.
\newblock
\urldef\tempurl%
\url{https://doi.org/10.1126/scirobotics.abm6074}
\showDOI{\tempurl}


\bibitem[Newcombe et~al\mbox{.}(2015)]%
        {newcombe2015aws}
\bibfield{author}{\bibinfo{person}{Chris Newcombe}, \bibinfo{person}{Tim Rath},
  \bibinfo{person}{Fan Zhang}, \bibinfo{person}{Bogdan Munteanu},
  \bibinfo{person}{Marc Brooker}, {and} \bibinfo{person}{Michael Deardeuff}.}
  \bibinfo{year}{2015}\natexlab{}.
\newblock \showarticletitle{How {Amazon Web Services} Uses Formal Methods}.
\newblock \bibinfo{journal}{\emph{Commun. ACM}} \bibinfo{volume}{58},
  \bibinfo{number}{4} (\bibinfo{year}{2015}), \bibinfo{pages}{66--73}.
\newblock


\bibitem[Newman et~al\mbox{.}(2022)]%
        {newman2022sigstore}
\bibfield{author}{\bibinfo{person}{Zachary Newman}, \bibinfo{person}{John~Speed
  Meyers}, {and} \bibinfo{person}{Santiago Torres-Arias}.}
  \bibinfo{year}{2022}\natexlab{}.
\newblock \showarticletitle{Sigstore: Software Signing for Everybody}. In
  \bibinfo{booktitle}{\emph{Proceedings of the 2022 {ACM} {SIGSAC} Conference
  on Computer and Communications Security ({CCS})}}.
  \bibinfo{pages}{2353--2367}.
\newblock


\bibitem[{Open Robotics}(2018)]%
        {ros2_lifecycle}
\bibfield{author}{\bibinfo{person}{{Open Robotics}}.}
  \bibinfo{year}{2018}\natexlab{}.
\newblock \bibinfo{title}{{ROS}~2 Managed Nodes (Lifecycle)}.
\newblock
  \bibinfo{howpublished}{\url{https://design.ros2.org/articles/node_lifecycle.html}}.
\newblock


\bibitem[{OpenAI}(2024)]%
        {openai2024functioncalling}
\bibfield{author}{\bibinfo{person}{{OpenAI}}.} \bibinfo{year}{2024}\natexlab{}.
\newblock \bibinfo{title}{Function Calling with {GPT}}.
\newblock \bibinfo{howpublished}{Vendor documentation,
  \url{https://platform.openai.com/docs/guides/function-calling}}.
\newblock


\bibitem[Papadimitriou(1979)]%
        {papadimitriou1979serializability}
\bibfield{author}{\bibinfo{person}{Christos~H. Papadimitriou}.}
  \bibinfo{year}{1979}\natexlab{}.
\newblock \showarticletitle{The Serializability of Concurrent Database
  Updates}.
\newblock \bibinfo{journal}{\emph{Journal of the {ACM}}} \bibinfo{volume}{26},
  \bibinfo{number}{4} (\bibinfo{year}{1979}), \bibinfo{pages}{631--653}.
\newblock


\bibitem[{Reproducible Builds Project}(2024)]%
        {reproducible_builds}
\bibfield{author}{\bibinfo{person}{{Reproducible Builds Project}}.}
  \bibinfo{year}{2024}\natexlab{}.
\newblock \bibinfo{title}{Reproducible Builds: a set of software development
  practices that create an independently-verifiable path from source to binary
  code}.
\newblock \bibinfo{howpublished}{\url{https://reproducible-builds.org/docs/}}.
\newblock


\bibitem[Sato(2014)]%
        {sato2014canary}
\bibfield{author}{\bibinfo{person}{Danilo Sato}.}
  \bibinfo{year}{2014}\natexlab{}.
\newblock \bibinfo{title}{Canary Release}.
\newblock
  \bibinfo{howpublished}{\url{https://martinfowler.com/bliki/CanaryRelease.html}}.
\newblock
\newblock
\shownote{Bliki post on martinfowler.com.}.


\bibitem[Schierman et~al\mbox{.}(2020)]%
        {schierman2020runtime}
\bibfield{author}{\bibinfo{person}{John~D. Schierman},
  \bibinfo{person}{Michael~D. DeVore}, \bibinfo{person}{Nathan~D. Richards},
  {and} \bibinfo{person}{Matthew~A. Clark}.} \bibinfo{year}{2020}\natexlab{}.
\newblock \showarticletitle{Runtime Assurance for Autonomous Aerospace
  Systems}.
\newblock \bibinfo{journal}{\emph{Journal of Guidance, Control, and Dynamics}}
  \bibinfo{volume}{43}, \bibinfo{number}{12} (\bibinfo{year}{2020}),
  \bibinfo{pages}{2205--2217}.
\newblock


\bibitem[Seto et~al\mbox{.}(1998)]%
        {seto1998simplex}
\bibfield{author}{\bibinfo{person}{Danbing Seto}, \bibinfo{person}{Bruce
  Krogh}, \bibinfo{person}{Lui Sha}, {and} \bibinfo{person}{Alongkrit
  Chutinan}.} \bibinfo{year}{1998}\natexlab{}.
\newblock \showarticletitle{The {Simplex} Architecture for Safe Online Control
  System Upgrades}. In \bibinfo{booktitle}{\emph{Proceedings of the {American}
  Control Conference ({ACC})}}. \bibinfo{pages}{3504--3508}.
\newblock


\bibitem[Sha(2001)]%
        {sha2001simplex}
\bibfield{author}{\bibinfo{person}{Lui Sha}.} \bibinfo{year}{2001}\natexlab{}.
\newblock \showarticletitle{Using Simplicity to Control Complexity}.
\newblock \bibinfo{journal}{\emph{{IEEE} Software}} \bibinfo{volume}{18},
  \bibinfo{number}{4} (\bibinfo{year}{2001}), \bibinfo{pages}{20--28}.
\newblock


\bibitem[{Spinnaker Community}(2023)]%
        {spinnaker_canary}
\bibfield{author}{\bibinfo{person}{{Spinnaker Community}}.}
  \bibinfo{year}{2023}\natexlab{}.
\newblock \bibinfo{title}{{Spinnaker} Automated Canary Analysis}.
\newblock
  \bibinfo{howpublished}{\url{https://spinnaker.io/docs/guides/user/canary/}}.
\newblock
\newblock
\shownote{Originally created at Netflix and Google.}.


\bibitem[Tang et~al\mbox{.}(2020)]%
        {tang2020twine}
\bibfield{author}{\bibinfo{person}{Chunqiang Tang}, \bibinfo{person}{Kenny Yu},
  \bibinfo{person}{Kaushik Veeraraghavan}, \bibinfo{person}{Jonathan Kaldor},
  \bibinfo{person}{Scott Michelson}, \bibinfo{person}{Thawan Kooburat},
  \bibinfo{person}{Aravind Anbudurai}, \bibinfo{person}{Matthew Clark},
  \bibinfo{person}{Kabir Gogia}, \bibinfo{person}{Long Cheng},
  \bibinfo{person}{Ben Christensen}, \bibinfo{person}{Alex Gartrell},
  \bibinfo{person}{Maxim Khutornenko}, \bibinfo{person}{Sachin Kulkarni},
  \bibinfo{person}{Marcin Pawlowski}, \bibinfo{person}{Tuomas Pelkonen},
  \bibinfo{person}{Andre Rodrigues}, \bibinfo{person}{Rounak Tibrewal},
  \bibinfo{person}{Vaishnavi Venkatesan}, {and} \bibinfo{person}{Peter Zhang}.}
  \bibinfo{year}{2020}\natexlab{}.
\newblock \showarticletitle{{Twine}: A Unified Cluster Management System for
  Shared Infrastructure}. In \bibinfo{booktitle}{\emph{14th {USENIX} Symposium
  on Operating Systems Design and Implementation ({OSDI})}}.
  \bibinfo{pages}{787--803}.
\newblock


\bibitem[Tang et~al\mbox{.}(2010)]%
        {tang2010overlap}
\bibfield{author}{\bibinfo{person}{Diane Tang}, \bibinfo{person}{Ashish
  Agarwal}, \bibinfo{person}{Deirdre O'Brien}, {and} \bibinfo{person}{Mike
  Meyer}.} \bibinfo{year}{2010}\natexlab{}.
\newblock \showarticletitle{Overlapping Experiment Infrastructure: More,
  Better, Faster Experimentation}. In \bibinfo{booktitle}{\emph{Proceedings of
  the 16th {ACM SIGKDD} International Conference on Knowledge Discovery and
  Data Mining ({KDD})}}. \bibinfo{pages}{17--26}.
\newblock


\bibitem[Todorov et~al\mbox{.}(2012)]%
        {todorov2012mujoco}
\bibfield{author}{\bibinfo{person}{Emanuel Todorov}, \bibinfo{person}{Tom
  Erez}, {and} \bibinfo{person}{Yuval Tassa}.} \bibinfo{year}{2012}\natexlab{}.
\newblock \showarticletitle{{MuJoCo}: A physics engine for model-based
  control}. In \bibinfo{booktitle}{\emph{{IROS}}}.
\newblock


\bibitem[Torres-Arias et~al\mbox{.}(2019)]%
        {torres2019intoto}
\bibfield{author}{\bibinfo{person}{Santiago Torres-Arias},
  \bibinfo{person}{Hammad Afzali}, \bibinfo{person}{Trishank~Karthik
  Kuppusamy}, \bibinfo{person}{Reza Curtmola}, {and} \bibinfo{person}{Justin
  Cappos}.} \bibinfo{year}{2019}\natexlab{}.
\newblock \showarticletitle{{in-toto}: Providing farm-to-table guarantees for
  bits and bytes}. In \bibinfo{booktitle}{\emph{{USENIX} Security Symposium}}.
\newblock


\bibitem[Verma et~al\mbox{.}(2015)]%
        {verma2015borg}
\bibfield{author}{\bibinfo{person}{Abhishek Verma}, \bibinfo{person}{Luis
  Pedrosa}, \bibinfo{person}{Madhukar Korupolu}, \bibinfo{person}{David
  Oppenheimer}, \bibinfo{person}{Eric Tune}, {and} \bibinfo{person}{John
  Wilkes}.} \bibinfo{year}{2015}\natexlab{}.
\newblock \showarticletitle{Large-scale cluster management at {Google} with
  {Borg}}. In \bibinfo{booktitle}{\emph{Proceedings of the Tenth European
  Conference on Computer Systems ({EuroSys})}}. \bibinfo{pages}{18:1--18:17}.
\newblock
\urldef\tempurl%
\url{https://doi.org/10.1145/2741948.2741964}
\showDOI{\tempurl}


\bibitem[Wang et~al\mbox{.}(2024)]%
        {wang2024voyager}
\bibfield{author}{\bibinfo{person}{Guanzhi Wang}, \bibinfo{person}{Yuqi Xie},
  \bibinfo{person}{Yunfan Jiang}, \bibinfo{person}{Ajay Mandlekar},
  \bibinfo{person}{Chaowei Xiao}, \bibinfo{person}{Yuke Zhu},
  \bibinfo{person}{Linxi Fan}, {and} \bibinfo{person}{Anima Anandkumar}.}
  \bibinfo{year}{2024}\natexlab{}.
\newblock \showarticletitle{{Voyager}: An Open-Ended Embodied Agent with Large
  Language Models}.
\newblock \bibinfo{journal}{\emph{Transactions on Machine Learning Research}}
  (\bibinfo{year}{2024}).
\newblock


\bibitem[Xue et~al\mbox{.}(2026a)]%
        {aeros_p4}
\bibfield{author}{\bibinfo{person}{Qin Xue} {et~al\mbox{.}}}
  \bibinfo{year}{2026}\natexlab{a}.
\newblock \bibinfo{title}{Governed Capability Evolution: Lifecycle-Time
  Compatibility Checking and Rollback for {AI}-Component-Based Systems, with
  Embodied Agents as Case Study}.
\newblock \bibinfo{howpublished}{arXiv:2604.08059}.
\newblock


\bibitem[Xue et~al\mbox{.}(2026b)]%
        {aeros_p2}
\bibfield{author}{\bibinfo{person}{Qin Xue} {et~al\mbox{.}}}
  \bibinfo{year}{2026}\natexlab{b}.
\newblock \bibinfo{title}{Learning Without Losing Identity: Capability
  Evolution for Embodied Agents}.
\newblock \bibinfo{howpublished}{arXiv:2604.07799}.
\newblock


\end{thebibliography}

\end{document}